\pdfoutput=1
\documentclass[11pt]{article}

\usepackage{emnlp2021}

\usepackage{times}
\usepackage{latexsym}

\usepackage[T1]{fontenc}
\usepackage[utf8]{inputenc}

\usepackage{times}
\usepackage{latexsym}
\usepackage{bm}
\usepackage{mathrsfs}
\usepackage{amsmath}
\usepackage{amsfonts}
\usepackage{graphicx}
\usepackage{amsthm}
\usepackage{hyperref}
\graphicspath{{./images/}}

\definecolor{lightgrey}{rgb}{0.925,0.925,0.925}
\definecolor{lightblue}{rgb}{0.905,0.995,1}
\definecolor{lightred}{rgb}{0.986,0.915,0.916}
\definecolor{blue}{rgb}{0.7,0.925,1}
\definecolor{red}{rgb}{0.986,0.85,0.86}
\definecolor{darkblue}{rgb}{0.5,0.725,1}
\definecolor{white}{rgb}{1,1,1}

\usepackage{microtype}

\title{WHOSe Heritage: Classification of {UNESCO} World Heritage Statements of ``{O}utstanding {U}niversal {V}alue'' with Soft Labels}

\author{Nan Bai$^{1}$, Renqian Luo$^{2}$, Pirouz Nourian$^3$, Ana Pereira Roders$^1$\\
  $^1$UNESCO Chair in Heritage and the Reshaping of Urban Conservation for Sustainability, \\
  Chair of Heritage and Values, Delft University of Technology, Delft, the Netherlands \\
  $^2$University of Science and Technology of China, Hefei, China \\
  $^3$Chair of Design Informatics, Delft University of Technology, Delft, the Netherlands \\
  \texttt{\{n.bai, p.nourian, a.r.pereira-roders\}@tudelft.nl}\\
  \texttt{lrq@mail.ustc.edu.cn}}

\begin{document}
\maketitle
\begin{abstract}
The UNESCO World Heritage List (WHL) includes the exceptionally valuable cultural and natural heritage to be preserved for mankind. 
Evaluating and justifying the Outstanding Universal Value (OUV) is essential for each site inscribed in the WHL, and yet a complex task, even for experts, since the selection criteria of OUV are not mutually exclusive.
Furthermore, manual annotation of heritage values and attributes from multi-source textual data, which is currently dominant in heritage studies, is knowledge-demanding and time-consuming, impeding systematic analysis of such authoritative documents in terms of their implications on heritage management.
This study applies state-of-the-art NLP models to build a classifier on a new dataset containing Statements of OUV, seeking an explainable and scalable automation tool to facilitate the nomination, evaluation, research, and monitoring processes of World Heritage sites.
Label smoothing is innovatively adapted to improve the model performance by adding prior inter-class relationship knowledge to generate soft labels. 
The study shows that the best models fine-tuned from BERT and ULMFiT can reach 94.3\% top-3 accuracy.
A human study with expert evaluation on the model prediction shows that the models are sufficiently generalizable.
The study is promising to be further developed and applied in heritage research and practice.\footnote{Code and data for this project are available at\\\href{https://github.com/zzbn12345/WHOSe_Heritage}{https://github.com/zzbn12345/WHOSe\_Heritage}}
\end{abstract}

\section{Introduction}
\label{sec:intro}

Since the World Heritage Convention was adopted in 1972, 1121 sites has been inscribed worldwide in the World Heritage List (WHL) up to 2019, aiming at a collective protection of the cultural and natural heritage of Outstanding Universal Value (OUV) for mankind as a whole \citep{UNESCO1972,VonDroste2011,PereiraRoders2011}. %UNESCO World Heritage Convention% 
First proposed in 1976, OUV, meaning the ``\emph{cultural and/or natural significance which is so exceptional as to transcend national boundaries and to be of common importance for present and future generations of all humanity}'', has been operationalized and formalized into an administrative requirement for new inscriptions on the WHL since 2005.
%become the administrative requirement for any nominations and inscriptions on the WHL since 2005
\citep{UNESCO2008,Jokilehto2006,Jokilehto2008}. %Operational Guides, What is OUV%
All nominations must meet one or more of the ten selection criteria (6 for culture and 4 for nature), focusing on different cultural and natural values.
%For example, \emph{Venice and its Lagoon}\footnote{http://whc.unesco.org/en/list/394} fulfills all six cultural criteria, while \emph{Sydney Opera House}\footnote{http://whc.unesco.org/en/list/103} indicates merely criterion (i), which stresses that it \emph{represents a masterpiece of human creative genius}.

Since 2007, complete \textbf{Statements of OUV} (SOUV) need to be submitted and approved for new World Heritage (WH) nominations, which should include, among others, a section of ``\emph{justification for criteria}'', giving a short paragraph to explain why a site (also known as property) satisfies each of the criteria it is inscribed under.
These statements are to be drafted by the State Parties after scientific research for any tentative nominations, further reviewed and revised by the Advisory Bodies from ICOMOS and/or IUCN, and eventually approved and adopted by the World Heritage Committee for inscription.
Similarly, Retrospective SOUV have been required for sites inscribed before 2006 to revise or refill the section \emph{justification of criteria} \citep{Retrospective2010}.
%Until the World Heritage Committee meeting in 2019, 78 Statements are still to be finalized and officially adopted.
However, the evaluation of SOUV can be ambiguous in the sense that: 
1) the selection criteria are not mutually exclusive and contain common information about historical and aesthetic/artistic values as an integral part \citep{Jokilehto2008};
2) the key stakeholders to evaluate the SOUV for a nomination occasionally disagree with each other at early stages, leading to recursive reviews and revisions, though all are considered to be domain experts \citep{Jokilehto2008,TarrafaSilva2010,VonDroste2011}.
A tool to check the accuracy, objectivity, consistency, and coherence of such statements can significantly benefit the inscription process involving thousands of experts worldwide each year.

Not only for new nominations, the SOUV are also essential reference points for monitoring and interpreting inscribed heritage sites \citep{Retrospective2010}.
Researchers and practitioners actively and regularly check if the justified criteria are still relevant for the sites, as to decide on further planning and managerial actions.
Moreover, these same statements are also used in support of legal court cases, should WH sites be endangered by human development \citep{PereiraRoders2010,VonDroste2011}.
Under the support of the \emph{Recommendation of Historic Urban Landscape} and the recent \emph{Our World Heritage} campaign, multiple data sources (e.g., news articles, policy documents, social media posts) are encouraged in such analyses of identifying and mapping OUV \citep{UNESCO2011,Bandarin2012,Ginzarly2019}.
%For public, visually showing why certain selection criteria have been justified with the WH property they visit can help them better understand and appreciate the cultural significance.
The traditional method of manually annotating heritage values and attributes by experts can be time-consuming and knowledge-demanding for analysing massive social media posts by people in cities with WH sites to find OUV-related statements, 
%and sometimes biased by a single expert's prior knowledge, 
albeit dominantly applied in practice \citep{TarrafaSilva2012,AbdelTawab2019,TarrafaSilva2010}.
%As such, it is difficult to scale up the analysis of heritage values and thus benefit from the massive multi-source data generate by various stakeholders worldwide \citep{Ginzarly2019}. %Ana Roders on WH, values%

To approximate both ultimate goals of this study: 1) aiding the inscription process by checking the coherence and consistency of SOUV, and 2) identifying heritage values from multiple data sources (e.g., social media posts), a computational solution rooted on SOUV is desired.
By training NLP models with the officially written and approved SOUV, a machine replica of the collective authoritarian view could be obtained.
This machine replica will not be employed at this stage to justify OUV for new nominations from scratch.
Rather, it will assess the written SOUV of WH sites (either existing or new) and classify OUV-related texts with the learned collective authoritarian view.
%Eventually, this machine replica will be employed in future studies to assess social media posts.
%Ideally, this machine would be capable of classifying a generic piece of text concerning OUV of WH, providing meaningful justifications and interpretations as if given by domain experts.
Furthermore, it can investigate the existing SOUV from bottom up and capture the subtle intrinsic associations within the statements and among the corresponding selection criteria \citep{bai2021}.
This 
%is a supplement to the current top-down process, yielding 
yields a new perspective on interpreting the WHL, which would give insights for furthering amending the concept of OUV and selection criteria to be better discernible.

%to automatically classify a piece of text harvest from either SOUV proposals or other more generic 

%A computational tool that can automate the heritage value identification could strongly enable authorities to study the effect of policies before implementation and stimulate the social inclusion in heritage management.
%Both needs point to NLP as solution, since: 
%1) NLP models, especially in the deep learning era, are built and trained from data as a bottom-up approach, giving chances for models to investigate the existing OUV document and induce the intrinsic associations among the written justification statements and the corresponding OUV criteria, which is a supplement to the current top-down process, yielding a new perspective on interpreting the WHL; 
%2) with the idea of transfer learning, the general language models pretrained on massive corpus and fine-tuned on domain-specific data are typically good at scaling up and generalizing to broader tasks, creating chances to further apply the trained OUV classifier on social media posts and policy documents which do not necessarily have an identical distribution with the training data \citep{Eisenstein2018,Rao2019,Jurafsky2020}. %general NLP textbooks%

Therefore, this study aims at training an explainable and scalable classifier that can reveal the intrinsic associations of World Heritage OUV selection criteria, which can be feasible to apply in real-world analyses by researchers and practitioners. 
%The classifier should have a reasonable performance, generalizability, resource utilization, and inference efficiency, making it feasible to apply in real-world analyses by researchers and practitioners. 
As outcome, this paper presents the classifier of UNESCO \underline{\textbf{W}}orld \underline{\textbf{H}}eritage Statements of \underline{\textbf{O}}UV with \underline{\textbf{S}}oft Lab\underline{\textbf{e}}ls (\textbf{WHOSe} Heritage).

The contributions of this Paper can be summarized as follows:
%\begin{itemize}
%\item 
1) A novel text classification dataset is presented, concerning a domain-specific task about Outstanding Universal Value for UNESCO World Heritage sites; 
%the dataset is formulated such that it contains characteristics of both multi-class and multi-label classification;
%\item 
2) Innovative variants of label smoothing are applied to introduce the prior knowledge of label association into training as soft labels,
%leverage the problem between multi-class and multi-label classification, 
which turned out effective to improve performance in most investigated popular models as baselines in this task;
%\item 
3) Several classifiers are trained and compared on the Statements of OUV classification task as initial benchmarks, supplemented with explorations on their explainability and generalizability using expert evaluation.
%This study can later benefit researchers and practitioners in real-world practice of heritage management, both for reviewing and revising Statements of OUV during the inscription process of new World Heritage properties, and for heritage studies and monitoring on existing ones.
%\end{itemize}

\section{Related Work}
\paragraph{Text classification}
%Text classification is an essential task in NLP. 
In the past decades, numerous models have been proposed from shallow to deep learning models for text classification tasks. 
In shallow learning models, the raw input text is pre-processed to extract features of the text, %such as Bag-of-words and N-gram, 
which are then fed
%, such as Bag-of-words (BOW), N-gram, term frequency-inverse document frequency (TF-IDF), word2vec~\citep{word2vec} and GloVe~\citep{glove}. 
%The features are then fed 
into machine learning classifiers, e.g., Naive Bayes~\citep{maron1961automatic} and support vector machine~\citep{joachims1998text} for prediction. 
In deep learning models, deep neural networks are leveraged to extract information from the input data, such as convolutional neural networks~(CNN)~\citep{kim2014convolutional,johnson2017deep}, recurrent neural networks~(RNN)~\citep{tai2015improved,cho2014properties}, attention networks~\citep{yang2016hierarchical} and Transformers~\citep{devlin-etal-2019-bert}.
Multi-class and multi-label tasks are two extensions of the simplest binary classification, where every sample can belong to one or more classes within a class list \citep{aly2005survey,tsoumakas2007multi}, where the labels may also be correlated \citep{pal2020magnet}.
This work explores the combined application of some popular shallow and deep learning models for a multi-class classification task.
%including both shallow and deep learning models.

%\paragraph{Multi-Class and Multi-Label Classification}
%Compared to binary classification, multi-class classifiers classify the input into one of several categories \citep{aly2005survey}. Multi-label classification is more complex than single-label classification, where one sample can belong to more than one class \citep{tsoumakas2007multi}. 
%A popular approach is binary relevance \citep{zhang2018binary}, which emulates the task into multiple binary classification problems which are easier to solve. 
%However, it ignores the correlation between the labels, which is evident in many tasks including the one in this work.
%Full consideration is therefore required on the semantic relationship among the labels. \citet{shimura2018hft} propose a hierarchical structure to leverage the hierarchical information of the multiple labels. 
%\citet{you2019attentionxml} incorporate attention to capture the information between the text and the labels. \citet{pal2020magnet} utilize graph attention networks to capture the correlations between labels, which is incorporated as prior knowledge.

\paragraph{Label Smoothing} Label smoothing (LS) is originally proposed as a regularization technique to alleviate overfitting in training deep neural networks \citep{szegedy2016rethinking,Muller2019}. 
It assigns a noise distribution on all the labels to prevent the model from predicting too confidently on `ground-truth' labels. It is widely used in computer vision \citep{szegedy2016rethinking}, speech \citep{chorowski2016towards} and natural language processing \citep{vaswani2017attention} tasks. 
Originally the distribution is uniform across the labels, which is data independent. 
Recently, other variants of LS are also proposed that are able to incorporate the interrelation information from the data into the distribution \citep{zhong2016towards,zhang2020delving,krothapalli2020adaptive}. 
In this work, the technique is applied to generate soft labels with a distribution derived from domain knowledge since the classes in this task are clearly interrelated with each other.

\paragraph{Transfer Learning in NLP} 
In many real-world applications, labelled data are limited and expensive to collect. 
Training models with limited data from scratch affects the performance. Transfer learning \citep{pan2009survey} is widely used to solve this by using word embeddings that are pretrained on massive corpus and fine-tuning them on target task. Earlier works~\citep{word2vec,glove} provide static word embeddings that ignore the contextual information in the sentences. More recent works, e.g., ULMFiT~\citep{ulmfit} and BERT~\citep{devlin-etal-2019-bert}, take the context into account and generate dynamic contextualized word vectors, showing excellent performance, which also prove to be sufficiently generalizable across many tasks.
%have been significant milestones in NLP.
This task, with a relatively small data size, employs the idea of transfer learning and applies both embedding methods.

\section{Data and Problem Statement}

\begin{table*}
\small
\centering
\begin{tabular}{lrrrrrrrrrrrr}
\hline \textbf{Split} & \textbf{C1} & \textbf{C2} & \textbf{C3} & \textbf{C4} & \textbf{C5} & \textbf{C6} & \textbf{N7} & \textbf{N8} & \textbf{N9} & \textbf{N10} & \textbf{Sum}\\ \hline
%meaning & masterpiece & influence & testimony & typology & land-use & associations & natural beauty & geology & evolution & biodiversity & - & -\\%
train & 333 & 631 & 651 & 774 & 209 & 327 & 386 & 261 & 370 & 572 & 4514\\
valid & 40 & 71 & 83 & 89 & 28 & 49 & 43 & 42 & 42 & 76 & 563\\
test & 41 & 79 & 72 & 92 & 35 & 47 & 45 & 32 & 50 & 71 & 564\\\hline
test in SD & 815 & 1563 & 1647 & 2049 & 554 & 876 & 510 & 334 & 465 & 548 & 9361\\
seen w LS & 1077 & 1747 & 1832 & 2131 & 609 & 1063 & 1130 & 630 & 1047 & 1251 & 12517\\
\hline
\end{tabular}
\caption{\label{dataset_description}
The number of samples in sentence level that contain each criterion as a label, annotated with C1 to C6 for cultural values and N7 to N10 for natural values. The first three rows show the data split using the field \emph{justification}; the fourth row shows a new dataset only for testing using the field \emph{short description} (SD); the last row shows the potential samples the models can see for each criterion after introducing label smoothing (LS).}
\end{table*}

\subsection{Data Collection and Pre-processing} 
\label{sec:data}
UNESCO World Heritage Centre openly releases a syndication dataset of the sites in XLS format\footnote{\href{http://whc.unesco.org/en/syndication}{http://whc.unesco.org/en/syndication}. Copyright © 1992 - 2021 UNESCO/World Heritage Centre. All rights reserved.}, which includes information of the inscribed World Heritage sites such as \emph{ID, name, short description}, \emph{justification of criteria} et. al..
%justified \emph{criteria}, 
Among them, the field of \emph{justification} provides a paragraph for each selection criterion the site fulfills\footnote{This field is not complete in the original XLS dataset. The WHC website is walked through to fill in the missing values.}, contributing as the input data for this task.
In total, 1052 out of 1121 WH sites contain the \emph{justification} data\footnote{The statistics are up to the 44th session of the World Heritage Committee held in Fuzhou, China in July 2021, after which the total number of WH sites grew to 1154.}, while the remaining 69 await the Retrospective SOUV to be approved as introduced in Section~\ref{sec:intro}.
As an example, in Venice and Its Lagoon, the paragraph on \textbf{criterion (i)} shows:
\begin{quote}
\small
    %\textbf{Criterion (i)}: The Sydney Opera House is a great architectural work of the 20th century. It represents multiple strands of creativity, both in architectural form and structural design, a great urban sculpture carefully set in a remarkable waterscape and a world famous iconic building.%
    %\textbf{Criterion (iii)}: 
    \emph{...The lagoon of Venice also has one of the highest concentrations of masterpieces in the world: from Torcello’s Cathedral to the church of Santa Maria della Salute.The years of the Republic’s extraordinary Golden Age are represented by monuments of incomparable beauty...\footnote{\href{https://whc.unesco.org/en/list/394}{https://whc.unesco.org/en/list/394}}}
\end{quote}
For any inscribed WH site $p_i \in P$, where $P$ is the set of all the sites, it may fulfill one or more of the ten selection criteria.
By checking if each criterion is justified for the site $p_i$, a non-negative vector $\boldsymbol{\gamma}_i := [\gamma_{i,k}]_{\kappa\times1},\ k\in[1,\kappa], \kappa=10$ can be formed as the ``parental'' label for the site:
\begin{equation}
\gamma_{i,k}=
    \begin{cases}
        1, & \text{if } p_i \ \text{meets the } k_{\text{th}} \ \text{criterion,}\\
        0, & \text{otherwise.}
    \end{cases}
\end{equation}

Meanwhile, the paragraphs $\boldsymbol{X}_i$ in the \emph{justification} field of $p_i$, describing all criteria that $p_i$ has, are split into sentences. 
For the $j_\text{th}$ sentence $\boldsymbol{x}_{i,j,k}$ describing the criterion $k$ possessed by the site $p_i$, a non-negative one-hot vector $\boldsymbol{y}_{i,j,k}$ can be formed as the ``ground-truth'' label for this single sentence:
\begin{equation}
    \boldsymbol{y}_{i,j,k}=\boldsymbol{e}_k\in\{0,1\}^\kappa.
\end{equation}
Each sentence $\boldsymbol{x}_{i,j,k} \in \boldsymbol{X}_i$ is treated as a sample, with two labels: a one-hot ``ground-truth label'' $\boldsymbol{y}_{i,j,k}$ for the particular sentence, and a multi-class ``parental label'' $\boldsymbol{\gamma}_i$ for all sentences that belong to the site $p_i$.
The sentence-level setup is desirable here since paragraphs may contain overwhelming information of multiple OUV criteria, as will be shown in Section~\ref{sec:assciation}. 
As such, a more specific indication of OUV tendencies in each part of the texts could be differentiated.
Complementarily, the fine-grained sentence-level prediction vectors could still be aggregated into paragraph/text levels without losing lower-level details, which will be demonstrated in Figure~\ref{bert-viz}.
As the sentences were written, revised, and approved by various domain experts at local and global levels during the inscription process, the labels can be considered as having a good ``inter-annotator agreement'' \citep{Jokilehto2008,Nowak2010}. %reference%

The following data pre-processing techniques are applied to construct the final dataset used for training: 
1) all letters are turned into lower-case; 
2) the umlauts and accents are normalized; 
3) numbers are replaced with a special $<\mathrm{NUM}>$ token; 
4) only sentences with a length between 8 and 64 words are kept, based on the dataset distribution; 
5) the sentences are randomly split into train/validation/test sets with a proportion of 8:1:1.
Additionally, the official definition sentences of selection criteria\footnote{\href{http://whc.unesco.org/en/criteria/}{http://whc.unesco.org/en/criteria/}} as given in Table~\ref{OUV_definition} of Appendix~\ref{selection_criteria} are respectively appended into the train split with the same one-hot sentence and parental labels for each criterion.
Stop-words are not removed since BERT and ULMFiT to be applied generally prefer natural texts with context information.
Furthermore, an additional $11_{\text{th}}$ class “Others” is introduced by appending an arbitrary noise of $\gamma_{i,\kappa+1}=0.2$ to all parental labels $\boldsymbol{\gamma}_i$, and a $0$ to all ``ground-truth'' labels $\boldsymbol{y}_{i,j,k}$, so that the models are not forced to give predictions only to the ten criteria even when the relevance to all of them is weak.
%to output a class related to none of the existing criteria, an 11th class is introduced by appending an arbitrary weight value $\gamma_{i,\kappa}=0.2$ to the end of all $\boldsymbol{\gamma}_i$ vectors and a $0$ to all $\boldsymbol{y}_{i,j,k}$ vectors.
For each sentence, the $11_{\text{th}}$ ``Others'' class and the complement sets of its parental labels could be regarded as the negative classes for classification since the site this sentence describes is not justified with those values.
An exemplary pre-processed data sample is shown in Table~\ref{sample} in Appendix~\ref{selection_criteria}.

On average, $27.97\pm11.04$ words appear in each sentence. A summary of the number of samples in sentence level in each split for each criterion is presented in the first three rows of Table~\ref{dataset_description}.

Similarly, the paragraphs $\boldsymbol{S}_i$ in the field \emph{short description} of WH site $p_i$, giving a general introduction of the site, which are not originally written to describe any specific OUV selection criterion, are pre-processed into an additional independent test dataset SD to evaluate the generalizability of the classifiers on unseen data that comes from a slightly different distribution. 
For those sentences $\boldsymbol{s}_{i,o} \in \boldsymbol{S}_i$, both ground-truth and parental labels are the same as $\boldsymbol{\gamma_i}$ for the site they describe. 
The total number of samples that contain each criterion in SD dataset is shown in the fourth row of Table~\ref{dataset_description}.

\subsection{Association between Classes}
\label{sec:assciation}
\citet{Jokilehto2008} summarized the selection criteria with their main focuses by inspecting the official definitions and the justification texts of WH sites. 
%Similarly, the topics for the remaining three can also be deduced from their definitions, as is shown in Table~\ref{OUV_topic}. 
Details about the definitions of the criteria could be found in Appendix \ref{selection_criteria}.
However, as stated in Section~\ref{sec:intro}, the criteria are not mutually exclusive. 
The \textbf{criterion (i)} justification of Venice in Section~\ref{sec:data} will be again used as an example.
Judging as a domain expert, it clearly describes criterion (i) as labelled, since it explicitly uses the term ``\emph{masterpieces}'' and ``\emph{monuments of incomparable beauty}''. 
However, traces can still be found on other values: 
 1) as it describes the ``\emph{Cathedral}'', ``\emph{church}'', and ``\emph{monuments}'', it also concerns the criterion (iv) about architectural \emph{typology}; 
 2) as it talks about the ``\emph{Golden Age}'', it also points to criterion (ii) about \emph{influence} and criterion (iii) about \emph{testimony}. 
In fact, Venice is also justified with criteria (ii), (iii), and (iv). 
Pragmatically speaking, for sites fulfilling more than one OUV selection criteria, it is hard to avoid talking about the other criteria while isolating one criterion alone \citep{PereiraRoders2010}.

Furthermore, the association between each pair of criteria can be different.
The distinction between criteria is generally larger when the pair comes from a different category (cultural v.s. natural).
For a pair of criteria from the same category, the association level can also vary. 
For example, \citet{Jokilehto2008} pointed out that ``\emph{criteria (i) and (ii) can reinforce each other while (iv) is often used as an alternative}''.
This complex association pattern can also be seen in the co-occurrence matrix $\boldsymbol{A_{\kappa\times\kappa}}:=[a_{k,l}]_{\kappa\times\kappa},\ k,l\in[1,\kappa]$ of the criteria in all the inscribed sites $P$, where the diagonal entries record the number of cases when each criterion is used alone (shown in Figure~\ref{co-occurrence_matrix} of Appendix~\ref{selection_criteria}):
\begin{equation}
a_{k,l}=
\begin{cases}
    \sum_i \left(\gamma_{i,k}\gamma_{i,l}\right), & \text{if } k\neq l,\\
    \sum_i \lfloor \frac{\gamma_{i,k}}{\sum_{j\in [1,\kappa]} \gamma_{i,j}} \rfloor, & \text{otherwise.}
\end{cases}
\end{equation}
This intrinsic association 
%implied by this co-occurrence pattern 
%(as shown in Figure~\ref{co-occurrence_matrix}) 
is to be used as the prior knowledge for the classification task.

\section{Models and Experiments}

\subsection{Soft Labels Generation}
\label{sec:label_smoothing}
Section~\ref{sec:assciation} argues that the selection criteria are not mutually exclusive, and that co-justified criteria of a WH site that have a stronger association may be reflected in the sentences describing a specific criterion.
In other words, classifying such sentences is not purely a single-label multi-class classification task.
Rather, it also has a multi-label characteristic considering the ``parental labels'' of the sites.

To leverage the problem between the two sorts of tasks and to prevent the models from being over-confident at the only ``ground-truth" labels, this paper proposes to apply the label smoothing (LS) technique with two novel variants to combine the ``ground-truth'' sentence label $\boldsymbol{y}_{i,j,k}$ and the parental document label $\boldsymbol{\gamma}_i$ into a single vector $\widetilde{\boldsymbol{y}}_{i,j,k}$ as soft labels for training process.
This is similar to the hierarchical LS approach proposed by \citep{zhong2016towards} to reflect the prior label similarity distribution.
We propose three variants: 
\textbf{vanilla} that assigns identical ``noises'' to all classes, which will be proved equivalent to the original LS in Appendix~\ref{proof}; 
\textbf{uniform} that treats all co-justified associated criteria in the parental label equally;
and \textbf{prior} that weights the co-justified criteria based on the frequency that the pair co-occurs in matrix $\boldsymbol{A}_{\kappa\times\kappa}$:
%The functions to compute $\boldsymbol{y}_{i,j,k}$ are shown below:

\begin{equation}
\label{eq_variant}
\centering
    \widetilde{\boldsymbol{y}}_{i,j,k} =
    \begin{cases}
        \mathbf{f}(\boldsymbol{y}_{i,j,k}+\alpha\mathbf{1}), & \text{if \textbf{vanilla}},\\
        \mathbf{f}(\boldsymbol{y}_{i,j,k}+\alpha\boldsymbol{\gamma}_i), & \text{if \textbf{uniform},}\\
        \mathbf{f}(\boldsymbol{y}_{i,j,k}+\alpha\boldsymbol{\mu}_k\odot\boldsymbol{\gamma}_i), & \text{if \textbf{prior}}.\\
    \end{cases}
\end{equation}
Here $\mathbf{f}:\mathbb{R}_{+}^d \to [0,1]^d$ is a variant of the original softmax function so that it maps a $d-$dimensional vector of non-negative real numbers to a distribution that sums up to $1$:
\begin{gather}
\label{eq_softmax}
    \mathbf{f}(\boldsymbol{z})_t = \frac{e^{z_t}-1}{\sum_{l=0}^d e^{z_l}-d}, \text{or } \mathbf{f}(\boldsymbol{z})=\frac{e^{\boldsymbol{z}}-\boldsymbol{1}}{{e^{\boldsymbol{z}^T}}\boldsymbol{1}-d},\\
    \text{for }t\in [0,d), \boldsymbol{1}:=[1]_{d\times1}\text{ and }\boldsymbol{z}:=[z_t]_{d\times1}\in\mathbb{R}_{+}^d;\nonumber
\end{gather}
$\alpha$ is a scalar that leverages the effect of LS; 
$\boldsymbol{\mu}_k:=[\mu_{l,k}]_{(\kappa+1)\times1}$ is a criterion-specific non-negative vector showing the inter-criteria associations:
\begin{equation}
    \mu_{l,k}=
    \frac{a_{l,k}}{\sum_i a_{i,k}}, \ l\in[1,\kappa+1],
\end{equation}
and $\odot$ represents the element-wise Hadamard-Schur product of vectors.
This variant of the softmax function introduced in Equation~\ref{eq_softmax} is preferable since it transforms the combined non-negative labels-vectors in Equation~\ref{eq_variant} to a “probability” distribution while keeping non-related labels still as 0.
For example, a combined vector $[2, 0, 1, 0]^T$ becomes $[.62, .08, .22, .08]^T$ with normal softmax, and $[.79, 0, .21, 0]^T$ with this variant.

All three variants are considered as options during training, and tuned as hyperparameters together with the scalar $\alpha\in\{0,0.01,0.05,0.1,0.2,0.5,1\}$. For all variants, the problem is purely multi-class when $\alpha=0$, and approaches multi-label when $\alpha$ gets larger, giving parental labels larger weights. 

The following benefits can be achieved with the use of proposed LS variants:
%\begin{itemize}
%\item 
1) The knowledge of the actual association of classes (selection criteria) are introduced into the training in both uniform and prior variants, giving the model chances to learn these intrinsic associations with soft labels;
%\item 
2) The freedom on the design decision of whether the problem should be multi-class or multi-label is provided for the model training process;
%\item 
3) The models can potentially see more instances for each class during training with LS variants, as shown in the last row of Table~\ref{dataset_description};
%\item 
4) The computed soft label vector $\widetilde{\boldsymbol{y}}_{i,j,k}$ is mathematically more similar to the prediction vector $\widehat{\boldsymbol{y}}_{i,j,k}$ than one-hot vectors, both of which are discrete ``probability'' distributions, pushing the use of Cross-entropy Loss closer to its original definition \citep{rubinstein2013cross}.
%\end{itemize}

\subsection{Baselines}
Five models are selected as baselines: 1) N-gram \citep{cavnar1994n} embedding followed by  multi-layer perceptron (MLP); 2) Bag-of-Embeddings (BoE) using GloVe \citep{glove}; 3) Gated Recurrent Unit (GRU) \citep{cho2014properties} with Attention \citep{bahdanau2014neural,yang2016hierarchical} (denoted as GRU+Attn); 4) Pretrained ULMFiT language model \citep{ulmfit} further fine-tuned on the full WHL domain dataset; and 5) uncased base BERT model \citep{devlin-etal-2019-bert}.
The former three models are trained mostly from scratch (where BoE and GRU+Attn used the GloVe-6B-300d vectors as initial embeddings), while the latter two are extensively pretrained and fine-tuned on this specific classification task.
The model implementation details and the hyperparameter configurations are shown in Appendix~\ref{model_detail}.

\subsection{Metrics}

For the training process, \textbf{Cross-Entropy} is used as the loss-function for two soft label vectors, while three metrics are used to evaluate the model performance as a multi-class classification task: 
1) \textbf{Top-1 Accuracy} which counts the instances when the predicted class with the highest output value matches the ground-truth sentence label; 
2) \textbf{Top-k Accuracy} which counts the instances when the ground-truth sentence label is among the top k predicted classes with the highest output values; 
3) \textbf{Macro-averaged F1} which calculates the overall cross-label performance.
\textbf{Per-class Metrics} (i.e., top-1 precision, recall, and F1) for each selection criteria are also calculated for evaluation purpose.

For the independent SD test set, two metrics are defined here to evaluate the model performance as a multi-label classification task:
1) \textbf{Top-1 Match} which counts the instances when at least one of the parental labels matches the predicted class;
2) \textbf{Top-k Match} which counts the instances when at least one parental label is among the top k predicted classes.
%2) \textbf{Top-k Jaccard-Index} which computes the size of intersection divided by the size of union of the top k predicted classes and the top k parental classes.
Arguably, the top-1 and top-k matches are more tolerant extensions of top-1 and top-k accuracy into multi-label classification scenarios.

For all evaluation metrics, k is chosen to be 3 following the rationale introduced in Appendix~\ref{selection_criteria}. 

\begin{table*}
\small
\centering
\begin{tabular}{lrrrrrrrrrr}
\hline \textbf{Model} & \textbf{LS Config} & \textbf{val 1} & \textbf{val k} & \textbf{val F1} & \textbf{test 1} & \textbf{test k} & \textbf{test F1} & \textbf{SD 1} & \textbf{SD k}\\ \hline
%\textbf{traditional}\\ \hline

N-gram & w/o LS & \colorbox{white}{67.38} & \colorbox{white}{90.82} & \colorbox{white}{63.11} & \colorbox{white}{59.96} & \colorbox{white}{88.87} & \colorbox{white}{58.87} & \colorbox{white}{70.49} & \colorbox{white}{95.13}\\
& uniform 0.1 & \colorbox{lightred}{67.19} & \colorbox{lightblue}{91.21} & \colorbox{red}{62.11} & \colorbox{lightred}{59.57} & \colorbox{blue}{89.65} & \colorbox{red}{58.24} & \colorbox{blue}{71.12}  & \colorbox{lightblue}{95.26} \\

BoE & w/o LS & \colorbox{white}{64.84} & \colorbox{white}{91.99} & \colorbox{white}{63.11} & \colorbox{white}{62.11} & \colorbox{white}{91.60} & \colorbox{white}{61.93} & \colorbox{white}{68.80} & \colorbox{white}{94.53}\\
& prior 0.01 & \colorbox{lightred}{64.26} & \colorbox{lightred}{91.60} & \colorbox{red}{62.48} & \colorbox{blue}{62.70} & \colorbox{lightred}{91.41} & \colorbox{lightblue}{62.14} & \colorbox{red}{66.15} & \colorbox{lightred}{94.14}\\

GRU+Attn & w/o LS & \colorbox{white}{64.26} & \colorbox{white}{91.60} & \colorbox{white}{60.83} & \colorbox{white}{60.55} & \colorbox{white}{91.41} & \colorbox{white}{59.28} & \colorbox{white}{64.27} & \colorbox{white}{92.71}\\
& uniform 0.2 & \colorbox{lightgrey}{64.26} & \colorbox{lightblue}{91.80} & \colorbox{blue}{61.36} & \colorbox{darkblue}{61.52} & \colorbox{red}{90.23} & \colorbox{darkblue}{61.06} & \colorbox{darkblue}{66.35}  & \colorbox{darkblue}{94.06} \\
\hline 
%\textbf{pre-trained}\\ \hline

ULMFiT & w/o LS & \colorbox{white}{69.34} & \colorbox{white}{93.95} & \colorbox{white}{68.40} & \colorbox{white}{66.41} & \colorbox{white}{92.38} & \colorbox{white}{66.09} & \colorbox{white}{70.21} & \colorbox{white}{96.15} \\
& prior 0.1 & \colorbox{blue}{70.12} & \colorbox{lightblue}{\textbf{\underline{94.34}}} & \colorbox{blue}{68.83} & \colorbox{darkblue}{67.19} & \colorbox{blue}{93.16} & \colorbox{blue}{66.97} & \colorbox{blue}{70.65} & \colorbox{lightgrey}{\textbf{\underline{96.22}}} \\

BERT & w/o LS & \colorbox{white}{70.31} & \colorbox{white}{\textbf{94.34}} & \colorbox{white}{69.60} & \colorbox{white}{\textbf{67.58}} & \colorbox{white}{93.55} & \colorbox{white}{67.15} & \colorbox{white}{\textbf{71.56}} & \colorbox{white}{95.96} \\
& uniform 0.2 & \colorbox{darkblue}{\textbf{\underline{71.68}}} & \colorbox{lightred}{93.95} & \colorbox{blue}{\textbf{\underline{70.42}}} & \colorbox{lightred}{66.99} & \colorbox{blue}{\textbf{\underline{94.53}}} & \colorbox{lightblue}{\textbf{\underline{67.34}}} & \colorbox{lightgrey}{71.51} & \colorbox{lightblue}{96.15}\\
\hline
\end{tabular}
\caption{\label{experiment_results}
The performance of models with and without LS on validation split, test split (top-1 accuracy, top-k accuracy, and averaged macro F1), and independent SD test set (top-1 match and top-k match), where k=3. The best score for each metric is highlighted in bold, and underlined if the best score occurs in models with LS. The effect of adding LS to each baseline is marked with background colors: blue indicates a rise in performance, red indicates a drop, while grey indicates a tie. The darker background color indicates a larger variation in performance.}
\end{table*}

\subsection{Experiment Setup}

The experiment consists of three successive steps for each baseline (details given in Appendix~\ref{model_detail}):
\begin{enumerate}
\item Grid search within a small range is performed to tune the hyperparameters with a single random seed, and the best configuration is selected according to the top-k accuracy on the validation split;
\item LS with different $\alpha$ values under all three conditions (vanilla, uniform, and prior) is tested using the configuration from step 1, repeated with 10 different random seeds, treated as another round of hyperparameter tuning, saving the best LS configuration according to the performance mean and variance over the seeds;
\item The best LS configuration in step 2 is applied to save a model with the same random seed used in step 1 and evaluated together with the baseline model without LS, both on validation/test splits and on SD test set;
%based on the aforementioned metrics, model size, and inference time;
\end{enumerate}
Early-stopping is applied during all training processes based on the top-k accuracy on the validation split. The models are implemented in PyTorch \citep{Rao2019} and experiments are performed on NVIDIA Tesla P100 GPU and Intel Core i7-8850H CPU, respectively.
The inference is performed entirely on a CPU to test the models' feasibility in more general application scenarios when GPU can be unavailable for end-users.
More details of training resource utilization, model size, and inference time is shown in Appendix~\ref{model_performance}.%NLP with Pytorch%

\section{Results and Analyses}

\subsection{Experiment Results}
\label{sec: result}

\begin{figure}
    \centering
    \includegraphics[width=\linewidth]{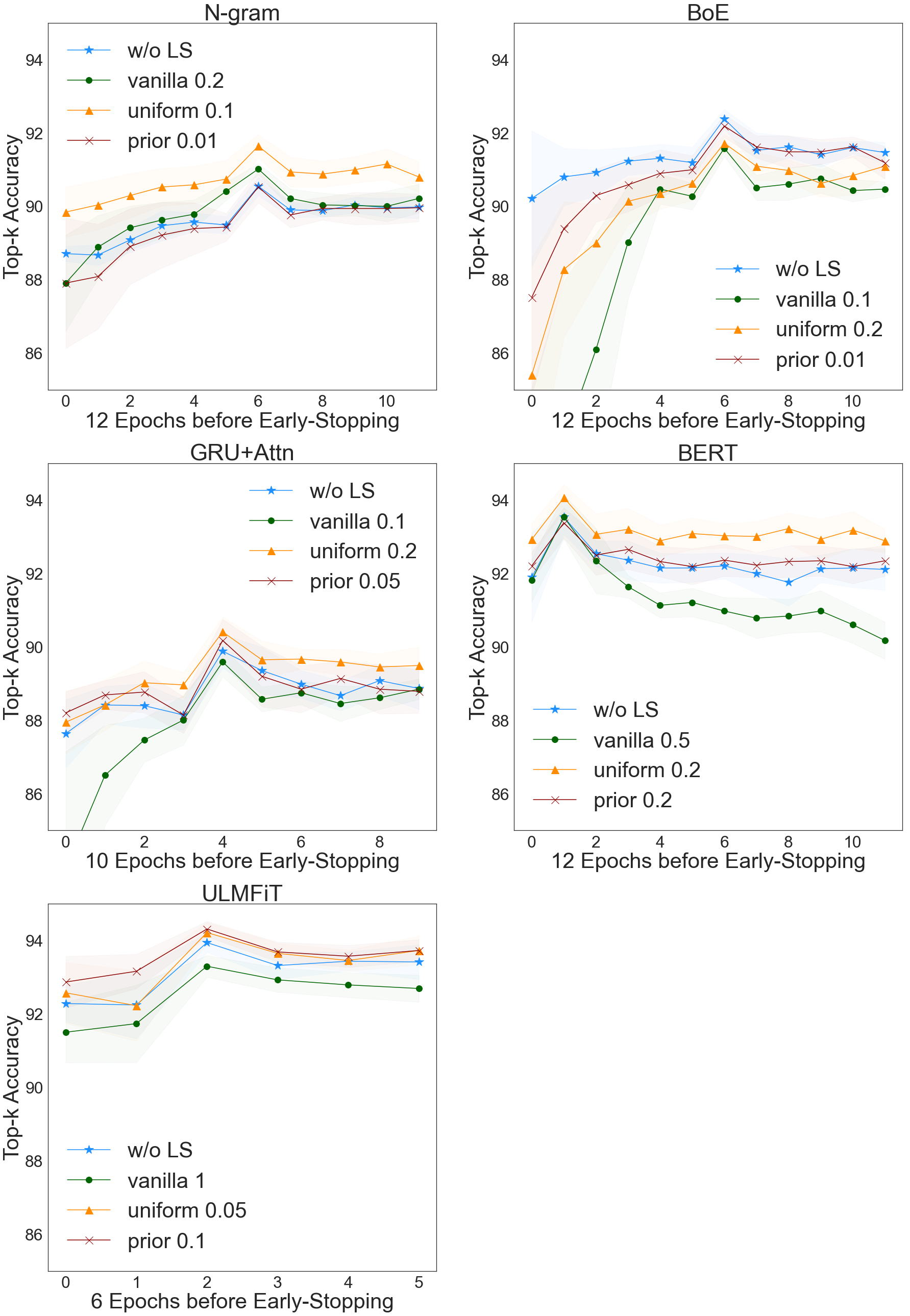}
    \caption{The average training curve of best-performing models in experiments under 10 random seeds for each baseline on validation split. The x-axes show several epochs before the early-stopping happened. The numbers of epochs are different for each baseline as described in Appendix~\ref{model_detail}. Orange curves with triangles show the top-k (k=3) accuracy with uniform LS, red curves with crosses the performance of prior LS, green curves with circles for vanilla LS, and blue curves with stars show the performance without LS. 95 \% confidence intervals of the performance based on the 10 random seeds are shown in shades.}
    \label{seeds}
\end{figure}

%\paragraph{General Effect} 
The averaged top-k accuracies of experiments conducted with 10 random seeds are shown in Figure~\ref{seeds}.
In most cases (except for BoE), 
%although the confidence intervals are sometimes overlapping, 
the models with proposed LS variants (uniform or prior) either strictly or weakly out-perform the baselines (without LS or with vanilla LS) based on multiple experiments.
Furthermore, the proposed LS variants seem to make the models more robust to over-fitting and catastrophic forgetting problems, especially with the cases of BERT and ULMFiT. 
The uniform variant of LS with different $\alpha$ values appears in most models.
A possible explanation is that uniform LS introduces the prior knowledge from the parental labels as ``noise'' in a simple way during the training, balancing yet not challenging the ``ground-truth'' sentence labels \citep{Muller2019}.
Yet, the complex effect of LS on different baselines invites further investigation.

%\paragraph{Effect of Smoothed Label}
Table~\ref{experiment_results} shows the performance of the models with and without LS on the validation split, test split, and SD test set. 
Except for BoE, introducing LS increased the performance of most baselines in most metrics.
Generally speaking, the pretrained models dominate the performance, and the highest score for all the metrics occurs in either ULMFiT or BERT, mostly with LS.
Still, top-1 accuracy only reaches 71\% in the best models, while top-k accuracy manages to reach 94\%, suggesting that it would be more reliable to look at the top 3 predictions during application in this task.
The models perform remarkably well in the SD test set, though given a relatively simpler task than in training, indicating the generalizability of the classifiers.

\begin{table}
\small
\centering
\begin{tabular}{llrrrr}
\hline \textbf{OUV} & \textbf{Focus} & \textbf{Prec} & \textbf{Recall} & \textbf{F1}\\ \hline
C1 & Masterpiece & 46.68 & 71.52 & 56.18\\
C2 & Values/Influences & 69.19 & 66.34 & 67.56\\
C3 & Testimony & 63.96 & 58.60 & 61.01\\
C4 & Typology & 61.10 & 54.23 & 57.24\\
C5 & Land-Use & 40.98 & 52.30 & 45.01\\
C6 & Associations & 58.28 & 67.89 & 61.27\\
N7 & Natural Beauty & 78.94 & 70.89 & 74.35\\
N8 & Geological Process & 66.92 & 80.42 & 72.39\\
N9 & Ecological Process & 60.16 & 67.23 & 63.45\\
N10 & Bio-diversity & 86.89 & 78.54 & 82.48\\
\hline
\end{tabular}
\caption{\label{per_class}
The average per-class metrics over all models on validation and test splits with LS, and the main focus of each criteria adapted from \citet{Jokilehto2008}.}
\end{table}

%\paragraph{Per-Class Performance}
The per-class top-1 metrics of the best models in each baseline on the validation and test split (Table~\ref{per_class}) make it evident that the difficulty for classifying each selection criterion varies.
$T$-test shows that F1 score is significantly different between the cultural and natural criteria ($t=8.20, p<.001$),
%(see Table~\ref{t-test}), 
suggesting that natural criteria are probably more clearly defined,
%and justified in practice
while cultural ones might be closely intertwined.
%Even within the same category, the performance on the criteria varies.
%Analysis of variance (one-way ANOVA) shows significant difference of F1 score among the classes both in cultural [$F(5,54)=12.32, p<.001$] and natural [$F(3,36)=26.10, p<.001$] categories.
%Post hoc comparisons using the Tukey HSD test indicate that criteria (v) and (ix) have significantly lower F1 than other cultural and natural OUV, respectively, and criterion (x) has a significantly higher F1.
The poor performance on criterion (v) is consistent with its smallest sample size (as shown in Table~\ref{dataset_description}); meanwhile, the models perform reasonably well for criterion (viii) with the second smallest sample size.
This suggests that except for sample size, the strong associations between the classes can also influence the difficulty for NLP models (and probably also for human experts) to distinguish the nuance of criteria.
Criterion (i) has a far poorer precision than recall, suggesting that samples from other criteria, especially from criterion (iv) based on the confusion matrices shown in Figure~\ref{confusion} of Appendix~\ref{model_performance}, are easily mistaken as this one.
This is also comprehensible since criterion (i), emphasizing that a site is a \emph{masterpiece}, can be easily mentioned ``unintentionally'' in the description of criterion (iv) that regards the value of some specific \emph{architectural typology}.

\subsection{Error Analysis and Explainability}
\label{sec:explainability}
%\paragraph{Lexicon-based Explanation}
%N-gram keywords with the highest prediction confidence for each class. \tbd

\begin{figure}
    \centering
    \includegraphics[width=\linewidth]{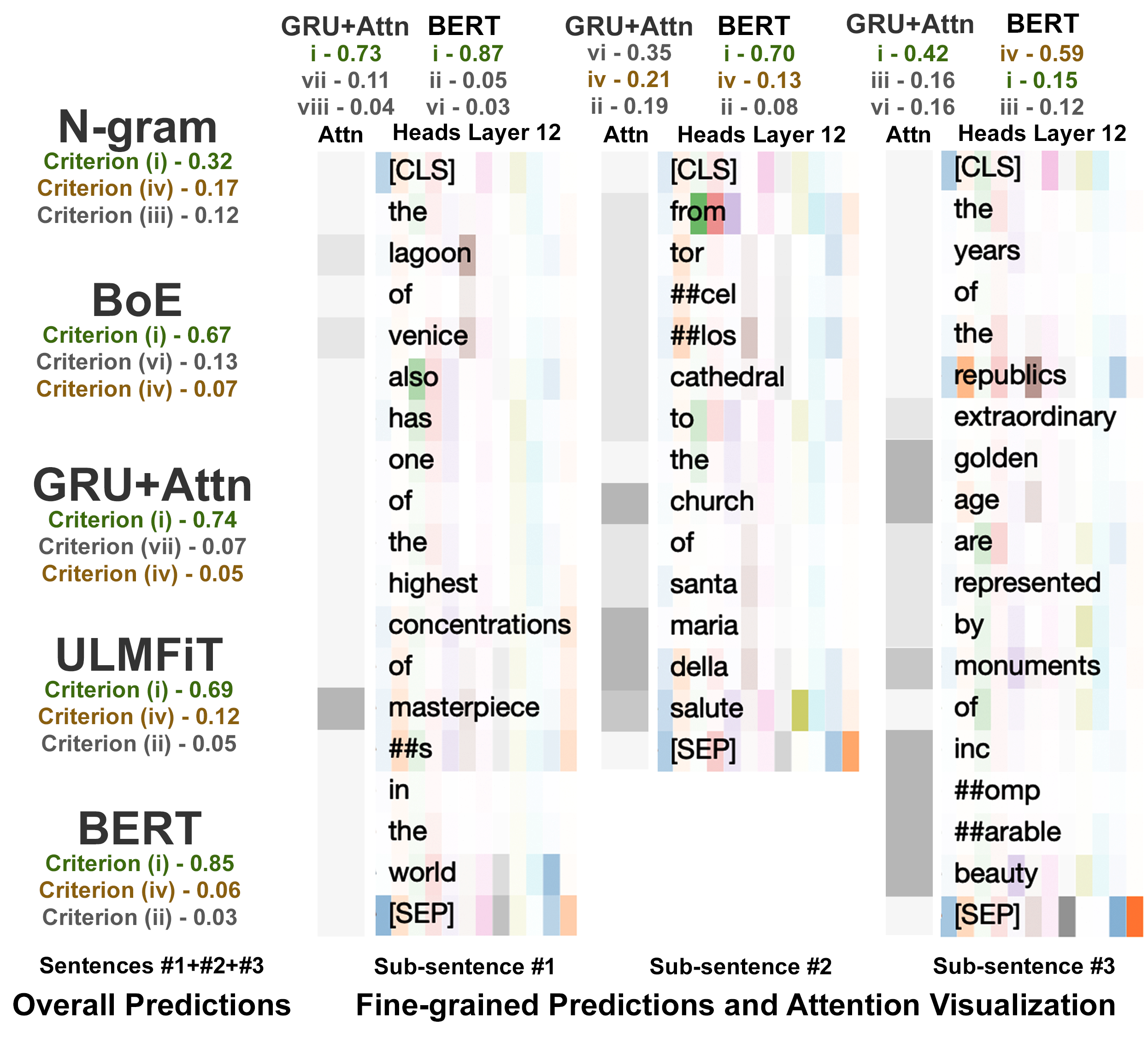}
    \caption{The overall and fine-grained top-3 predictions of models, and attention weights of GRU+Attn and BERT models on the exemplary sub-sentences concerning criterion (i) in Venice. The left part of the image reports the top-3 predictions of all 5 models when the models take the aggregated paragraph as input. The top part reports the fine-grained top-3 predictions of two models on each sub-sentence. The rest of the image visualizes the attention weights. Attention weights of GRU+Attn is visualized in grey-scale, and that of BERT is illustrated using BertViz as coloured bars.}
    \label{bert-viz}
\end{figure}
%\paragraph{Embedding-based Explanation}
%Expansion of GloVe Embedding space after fine-tuning. \tbd

%\paragraph{Attention-based Explanation}
%Highlighting the attended words in GRU and BERT heads. \tbd

Although sometimes challenged \citep{Serrano2020}, attention mechanisms are believed to be effective for visualizing NLP model performance in an explainable manner \citep{yang2016hierarchical,vaswani2017attention, Tang2019, Sun2020}. 
The same example on OUV selection criterion (i) in Venice as in Section \ref{sec:data} and \ref{sec:assciation} will be demonstrated here using the trained models from the attention-enabled GRU+Attn and BERT, as shown in Figure~\ref{bert-viz}, with the help of BertViz library \citep{vig-2019-multiscale,tensor2tensor}.
GRU+Attn employs a single universal attention mechanism to all inputs, while BERT has 12 attention heads for the [CLS] token on its last layer, both of which manage to capture the meaningful keywords and phrases such as \emph{masterpiece}, \emph{church}, \emph{golden age}, \emph{monuments}, and \emph{incomparable beauty} in the sentences. As a note, \citet{clark-etal-2019-bert} used probing to find out that some BERT attention heads correspond to certain linguistic phenomena.
In this study, the attention heads from the last layer also seem to focus on different semantic information of OUV.
This observation invites further studies.

Figure~\ref{bert-viz} also shows the top-3 predictions of the models on the exemplary sentences.
In the overall predictions taking the sentences as a paragraph for input, all models manage to give the ground-truth label criterion (i) the highest predicted value (from 0.32 in N-gram to 0.85 in BERT).
Remarkably, all models also include criterion (iv) in the top-3 predictions (from 0.05 in GRU+Attn to 0.17 in N-gram), suggesting that the sentences might also be related to criterion (iv).
The fine-grained predictions taking each sub-sentence as input, however, show a different pattern.
Although criterion (i) is almost always present in the top-3 predictions, criterion (iv) shows to take a higher place in the second sentence by GRU+Attn, and in the third sentence by BERT.
This behaviour is not necessarily an error per se in prediction. 
Rather, considering the arguments in Section \ref{sec:assciation}, those sub-sentences could be indeed relevant to other criteria (in this case, criterion iv) based on the association pattern, q.v. \citet{bai2021}, indicating why criterion (iv) is always included in the overall predictions.
%The performance demonstrated here shows the association and co-occurrence of criteria (i) and (iv) in the example, consistent with the argumentation in Section~\ref{sec:assciation}.
%while BERT seems to be more confident with the result without losing the association information between the criteria.

\begin{figure}
    \centering
    \includegraphics[width=\linewidth]{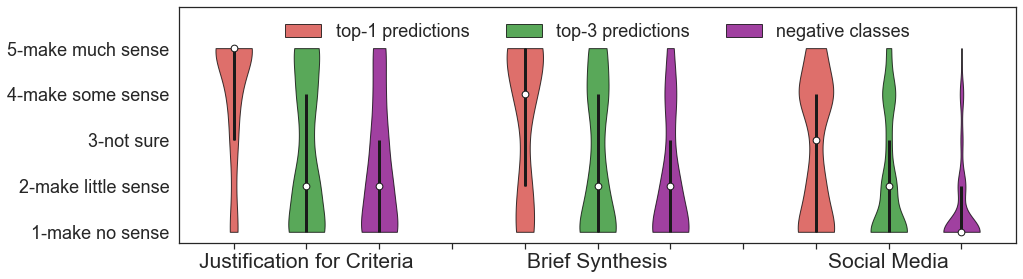}
    \caption{The distribution as violin plots of expert evaluations given to the relevance of selection criteria and sample sentences about Venice from three sources. The scores for top-1 and top-3 classes and the negative class predicted by the models are plotted separately. The 25\%, 75\% percentiles, and the medians are also shown.
    }
    \label{human}
\end{figure}

%\begin{table}
%\small
%\centering
%\begin{tabular}{lrrrrr}
%\hline \textbf{Metric} & \textbf{$t$ value} &  \textbf{$p$ value} & \textbf{$n_\text{C}$} & \textbf{$n_\text{N}$} &\textbf{Cohen's $d$}\\ \hline
%Precision & 6.36* & $<$.001 &60 & 40& 1.30\\
%Recall & 6.14* & $<$.001 &60 & 40& 1.25\\
%F1 & 8.20* & $<$.001 &60 & 40& 1.67\\
%\hline
%\multicolumn{6}{l}{\small *$p<.001$ with large effect size.}
%\end{tabular}
%\caption{\label{t-test}
%The result of independent sample $t$-test on the per-class metrics between cultural and natural OUV of all models with LS on the test and validation splits.}
%\end{table}

\subsection{Expert Evaluation}
\label{sec:human}

Eight heritage researchers with rich experience in identifying heritage values and attributes were invited for a human study adapted from \citet{he2021read}, \citet{nguyen2018comparing} and \citet{schuff2020explainable}, to test the models' reliability and generalizability.
They were presented with 56 sentences about Venice harvested from ``\emph{Justification}'' (14) and ``\emph{Brief Synthesis}'' (13) in SOUV and Social Media platforms (29).
Each sentence was given three positive classes as top-1 and top-3 criteria predictions from BERT and ULMFiT models, and one negative class as another random cultural criterion.
Not knowing that the criteria are predictions by computer models, the experts were asked to rate the relevance of the sentences and each criterion on a 5-point Likert scale.

The distributions of all the ratings are shown in Figure~\ref{human}.
For all data sources, the expert ratings for top-1 and top-3 predictions are significantly higher than those for negative classes based on Mann-Whitney $U$ tests (See Table~\ref{u-test} in Appendix~\ref{expert_evaluation}).
The average ratings of experts for each sentence-criterion pair show a strong correlation with the average confidence scores of models ($r=0.618, p<0.001$).
Some heritage experts seem to be rather cautious and reserved to assess informal texts as ``culturally significant'' without further historical contexts and comparative studies.
%the informal texts from social media, giving lower ratings than formal SOUV texts.
For example, the third sentence in Table~\ref{example} of Appendix~\ref{expert_evaluation} from social media, ``\emph{In 1952, the station was finalized on a design by the architect Paul Perilli}'' with a predicted label of criterion (i) got extremely divergent expert scores.
For some experts, it is clearly related to criterion (i) about \emph{masterpiece} based on the semantic content.
However, for the experts who rated a low score, merely declaring that some building is designed by a certain architect does not automatically entail that it is a masterpiece.
Further investigations have to be made to fully convince them.
%, compared to the formal SOUV texts. 
Although such an example shows disagreement amongst the experts and between the experts and the computer models, it does not limit the machine's ability to differentiate positive and negative classes.
%even when fed by informal texts harvested from social media.
%Arguably, the lower ratings on social media data do not necessarily indicate failure of generalization of the model.
%Rather, many disagreements among the experts and between the experts and models are due to the careful and critical standards on justifying OUV by heritage scholars, especially when the language is informal.
%, showing the complexity of human justification.
Full details of the human study are presented in Appendix~\ref{expert_evaluation}.
The expert evaluation proves that the models are sufficiently reliable and capable of identifying OUV-related statements even from the less formal social media data, useful for the ultimate motivations of this study discussed in Section~\ref{sec:intro}.

\section{Discussion and Conclusions}
\label{sec:discussion}

%This study shows that the introduction of the soft labels has weakly improved the performance of most of the classifiers and that the difference made is nearly significant statistically. This invites further research into the applicability of such an approach. However, it must be noted that as of yet the input labels have not been independently assigned by other humans or machines and as such the contribution of the work is first and foremost the development a first benchmark data-set for classification of this specific corpus and the making of a first workable data-set for further specific NLP studies in the area of OUV.

This paper presents a new text classification benchmark from a real-world problem about UNESCO World Heritage Statements of Outstanding Universal Value (OUV).
The problem is essentially a multi-class single label classification task, while the classes are not necessarily mutually exclusive.
The prior knowledge of the class association is added to the training process as soft labels through novel variants of label smoothing (LS).
The study shows that introducing LS improved the performance on most baselines,
%including the pretrained language models ULMFiT and BERT, 
reaching a top-3 accuracy of 94.3\%.
The models also performed reasonably well in an independent test dataset and received positive outcomes in a human study with domain experts, suggesting that the classifiers have the potential to be further developed and applied in the World Heritage research and practice.

LS was not tuned together with other hyperparameters during the training. Yet, it still showed an improvement in most baselines.
However, the complex effect of LS on different baselines needs more investigation.
The top-1 accuracy is limited even on the best models, which is not uncommon in the literature for non-binary multi-class classification when the labels are not sufficiently distinct \citep{Sun2019}.
Applying data augmentation and training supplemental binary classifiers may improve the performance on difficult classes.
The choice of replacing all numbers into $<\mathrm{NUM}>$ tokens might introduce both advantages and drawbacks in terms of semantic context and generalizability when historical dates might be crucial information, which invites more investigations.
Moreover, more studies on the generalizability and reliability of the models on data from different distributions (e.g., from policy documents or news article) are needed before further application.
This work would support a series of follow-up studies respectively exploring the intrinsic associations of OUV based on the models’ behaviour~\citep{bai2021}, application of the proposed methods in social media mining in Venice~\citep{bai2021_2}, and generalizability in case studies worldwide.

%The current OUV evaluation and justification process is time-consuming and knowledge-demanding.
%By innovatively introducing NLP into the field of heritage studies, chances are created to comprehensively investigate the semantic meaning of OUV selection criteria and the intrinsic associations among them.
%As argued in Section~\ref{sec: result}, natural OUV criteria may have a clearer definition than cultural ones.
%This might lead to another round of discussion and even revision of the OUV definitions \citep{Jokilehto2008}, which is explicitly suggested by World Heritage Committee (WHC) to reflect the evolution of WH concept itself\footnote{http://whc.unesco.org/en/criteria/}.

This work is intended to aid, but not replace the workload of human stakeholders: for State Parties to identify OUV-related statements through documentation, for Advisory Bodies and WHC to review and revise the yearly nomination proposals, for researchers to investigate massive official discourse and user-generated content, and for the public to visually understand the values of \emph{Their World Heritage} around them.
Therefore, this work \textbf{WHOSe Heritage} can be another milestone for the digital transformation of World Heritage studies, aiming at a more socially inclusive future practice.

%Limitation and future work. Further transferability and generalizability to data with probably different distribution. Further explanation on the classes and suggestions for UNESCO World Heritage Centre on OUV. \tbd

%The social relevance and contribution of AI techniques, NLP to be specific, to traditional field of heritage conservation in the new era. \tbd

\section*{Acknowledgements}
The presented study is within the framework of the Heriland-Consortium.
HERILAND is funded by the European Union’s Horizon 2020 research and innovation programme under the Marie Sklodowska-Curie grant agreement No 813883.
The authors are grateful for all the constructive comments from the anonymous reviewers.

\section*{Broader Impact Statement}
This work focuses on exploring and applying NLP techniques to a real-world application of cultural and natural World Heritage (WH) preservation for the sake of social good. 
The research is to aid the identification and justification of heritage values across the world for various stakeholders, including both heritage experts and lay-persons, through text classification, as is pointed out in Section~\ref{sec:intro} and \ref{sec:discussion}.
It can lead to better understandings of the OUV criteria and the association among them.
%, yielding chances to eventually amend and clarify some OUV definition which may be ambiguous.

The dataset used in this work is collected by the author(s) from the public website of UNESCO World Heritage Centre via XLS syndication respecting the terms of use and copy rights. 
The description of the dataset is sufficiently revealed in section~\ref{sec:data} and Appendix~\ref{selection_criteria}.
All labels used are based on the official OUV justification given by local and global heritage experts and involve no crowd workers or other new annotators.
The dataset and the methods used in the paper do not contain demographic/identity characteristics.
Once deployed, the model does not learn from user inputs, and it generates no harmful output to users.
The expert evaluation involving human study was totally voluntary, did not collect any personal information, and the privacy of the experts was fully protected.
Though initially unaware of the true purpose of the evaluation to reduce bias, the experts were explained with the study afterwards.

BERT and ULMFiT with LS proved to perform best in all investigated metrics. However, there is a trade-off to consider for real-world application. As claimed in Appendix~\ref{model_performance} and Section~\ref{sec:explainability}, ULMFiT has a relatively shorter inference time compared to BERT, while BERT is potentially more explainable due to the attention mechanism. Both models might work optimally for different application scenarios.

Nevertheless, the interpretation of the classification result needs to be carefully conducted by researchers and practitioners, especially during policy decision-making on World Heritage for the social benefit of the entire human species.
WH inscription and OUV justification are far more complicated than only reading written texts and identifying the described values.
Rather, it is a systematic thematic study based on scientific research and always rooted in a COMPARATIVE study across the globe \citep{Jokilehto2008}.
The actual decisions of including new nominations into the WHL have to be made by human with heritage investigations.
This is also evident in the results of expert evaluation and during the open discussion about the exercise with invited experts.
As stated in the example shown in Section~\ref{sec:explainability}, thorough heritage investigations are always needed to determine if a site truly justifies certain OUV selection criteria.
Such investigations, however, would be out of the scope of our NLP application study investigating the semantic and syntactic content of written official documents.
Therefore, a human has to be involved in the loop during application.

This study and the obtained NLP models are inherently less biased than manual annotation by a single expert in the sense that they avoid adding too much implicit personal experience into the written texts, and that the trained models represent the collective views of many human experts in the past.
This can also be seen in some divergent evaluation outcomes by the eight invited experts, as demonstrated in Appendix~\ref{expert_evaluation}: though one specific expert may be more cautious and critical at a certain sample, the overall trend of all experts can consistently differentiate the positive and negative classes.
However, the computational models trained on SOUV can also be a double-edged sword in the sense that they are highly dependent on the existing descriptions, which may contain historical unfairness.
%and do not necessarily contribute to identify some ``universally outstanding value'' compared to all other WH candidates.

Researchers and practitioners, especially those outside of the Computer Science field, need to be explicitly informed and even warned before usage on the limitations of such models, to avoid automation bias, which shows that people favour the results automatically generated from systems for decision-making \citep{Automation2010}.
%Failure of the proposed method/system and misinterpretation of the results can lead to severe negative impact.
Wrongly under-judging the value of a WH nomination merely based on text classification results and consequently deferring or even refusing the inscription can cause a great loss to human culture in the worst scenario, as it can hamper its access to the available heritage management and conservation programs.
Therefore, this work functions as a supplemental tool and reference for the understanding/evaluating of World Heritage OUV implied in text descriptions, which will and shall not replace the human effort and/or deviate the expert knowledge in WH decision-making process.
Instead, it has two ultimate goals as use-cases: 1) aiding inscription processes by checking the coherence and/or consistency of OUV statements; 2) mining heritage-values-related texts from multiple data sources (e.g., social media).

\bibliography{anthology,custom}
\bibliographystyle{acl_natbib}

%\appendix

%\section{Example Appendix}
%\label{sec:appendix}

%This is an appendix.
\appendix
\section*{Appendix}
\section{Selection Criteria and Dataset}
\label{selection_criteria}

\begin{figure}[htbp]
    \centering
    \includegraphics[width=0.9\linewidth]{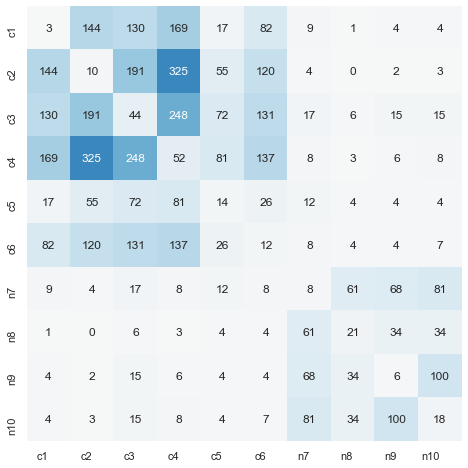}
    \caption{The co-occurrence matrix $\boldsymbol{A}_{\kappa\times\kappa}$ of OUV selection criteria in all UNESCO WH sites.}
    \label{co-occurrence_matrix}
\end{figure}

\begin{table*}
\centering
\small
\begin{tabular}{lllr}
\hline \textbf{OUV} & \textbf{Focus} & \textbf{Definition} & \textbf{Total} \\ \hline
%meaning & masterpiece & influence & testimony & typology & land-use & associations & natural beauty & geology & evolution & biodiversity & - & -\\%
C1 & Masterpiece & \emph{To represent a masterpiece of human creative genius}; & 254\\
C2 & Values &\emph{To exhibit an important interchange of human values, over a span of time or within} & 449\\
&/Influences&\emph{a cultural area of the world, on developments in architecture or technology,}\\
&& \emph{monumental arts, town-planning or landscape design};\\
C3 & Testimony &\emph{To bear a unique or at least exceptional testimony to a cultural tradition or to a} & 466\\
&&\emph{civilization which is living or which has disappeared};\\
C4 & Typology &\emph{To be an outstanding example of a type of building, architectural or technological}& 597\\
&& \emph{ensemble or landscape which illustrates (a) significant stage(s) in human history};\\
C5 & Land-Use &\emph{To be an outstanding example of a traditional human settlement, land-use, or}& 157\\
&& \emph{sea-use which is representative of a culture (or cultures), or human interaction with} \\
&&\emph{the environment especially when it has become vulnerable under the impact of}\\
&& \emph{irreversible change};\\
C6 & Associations &\emph{To be directly or tangibly associated with events or living traditions, with ideas, or} & 246\\
&& \emph{with beliefs, with artistic and literary works of outstanding universal significance};\\
N7 & Natural &\emph{To contain superlative natural phenomena or areas of exceptional natural beauty} & 146\\
& Beauty&\emph{and aesthetic importance};\\
N8 & Geological&\emph{To be outstanding examples representing major stages of earth's history, including} & 93\\
&Process &\emph{the record of life, significant on-going geological processes in the development of} \\
&&\emph{landforms, or significant geomorphic or physiographic features};\\
N9 & Ecological &\emph{To be outstanding examples representing significant on-going ecological and} & 128\\
&Process &\emph{biological processes in the evolution and development of terrestrial, fresh water,} \\
&&\emph{coastal and marine ecosystems and communities of plants and animals};\\
N10 & Bio-diversity &\emph{To contain the most important and significant natural habitats for in-situ conservation} & 156\\
&&\emph{of biological diversity, including those containing threatened species of outstanding} \\
&&\emph{universal value from the point of view of science or conservation}.\\\hline
\end{tabular}
\caption{\label{OUV_definition}
The definition for each UNESCO World Heritage OUV selection criterion and its main topic according to \citet{UNESCO2008}, \citet{Jokilehto2008}, and \citet{bai2021}. The last column shows the total number a criterion is justified with a WH site either uniquely or together with other criteria until 2019.}
\end{table*}

\begin{table}
\centering
\small
\begin{tabular}{rrrl}
\hline \textbf{N} & \textbf{Count} & \textbf{Proportion} & \textbf{Example} \\ \hline
%meaning & masterpiece & influence & testimony & typology & land-use & associations & natural beauty & geology & evolution & biodiversity & - & -\\%
1 & 188 & 16.75\% & Sydney Opera House\\
2 & 468 & 41.71\% & Babylon\\
3 & 304 & 27,09\% & City of Bath\\
4 & 103 & 9.18\% & Yellowstone National Park\\
5 & 34 & 3.0\% & Acropolis, Athens\\
6 & 4 & 0.36\% & Venice and its Lagoon\\
7 & 2 & 0.18\% & Mount Taishan\\
%sum & 1122 & 100\%\\
\hline
\end{tabular}
\caption{\label{co-justified}
The distribution of the total number of selection criteria $\sum_{k=1}^\kappa{\gamma}_{i,k}$ a site is justified with.}
\end{table}

\paragraph{Selection Criteria Definitions}
%As explained in \citet{UNESCO2008}, 
For any site to be inscribed in the World Heritage List, it must satisfy at least one of the ten Outstanding Universal Value (OUV) selection criteria and meet the conditions of integrity and/or authenticity.
%\citep{Jokilehto2008}.
%\citet{Jokilehto2008} summarize the six cultural criteria and the first natural criterion with their essential topics, which can also be completed for the last three criteria, as is shown in Table~\ref{OUV_topic}.

However, it is to be stressed that the definition of the selection criteria shown in Table~\ref{OUV_definition} is regularly revised by the World Heritage Committee to reflect the evolution of World Heritage (WH) itself\footnote{\href{http://whc.unesco.org/en/criteria/}{http://whc.unesco.org/en/criteria/}}.
For example, cultural (criteria i-vi, also denoted as C1-C6) and natural (criteria vii-x, also denoted as N7-N10) OUV used to be justified apart as two sets.
Since 2004, the two sets are combined.
Although WH sites are usually justified with OUV from one category (cultural or natural), within the domain of mix heritage and cultural landscape, OUV from both categories can co-occur in one site (e.g., Mount Tai has all first seven criteria).

\paragraph{Association between Criteria}
Among all the 1121 sites inscribed in the World Heritage List up to 2019, only 188 are justified with only one criterion.
The distribution of the total number of criteria justified for each site (i.e., $\sum_{k=1}^\kappa{\gamma}_{i,k}$) is shown in Table~\ref{co-justified}.
This is an indication on the extend of how the problem characterizes a multi-label classification nature.
It is also the rationale behind the choice of $\text{k}=3$ for the evaluation metrics Top-k Accuracy and Top-k Match, as 85.5\% of sites are justified with no more than 3 criteria.
Regardless of the number of co-justified criteria for each site, the co-occurrence matrix $\boldsymbol{A}_{\kappa\times\kappa}$ of all selection criteria is shown in Figure~\ref{co-occurrence_matrix}.
The row-normalized $\boldsymbol{A}_{\kappa\times\kappa}$ becomes the source of the criterion-specific non-negative vectors $\boldsymbol{\mu}_k$ of the \textbf{prior} variant of Label Smoothing (LS), as is discussed in Section~\ref{sec:label_smoothing}.
The criteria from the same category are co-justified more often, while criteria (ii-iv), (iii-iv), and (ii-iii) are the most frequently co-occurred pairs.
%, which is different from the observation of \citet{Jokilehto2008} approximately a decade ago.

\begin{table}
\centering
\small
\begin{tabular}{lrl}
\hline \textbf{Attribute} & \textbf{Symbol} & \textbf{Data} \\ \hline
data & $\boldsymbol{x}_{i,j,k}$ & \emph{the counter reformation of}\\
&&\emph{the late $<\mathrm{NUM}>$ th}\\
&& \emph{century led to a flowering in}\\
&&\emph{the creation of calvaries}\\
&&\emph{in europe}\\
single label & $k$ & Criterion (iv)\\
sentence label & $\boldsymbol{y}_{i,j,k}$ & [0, 0, 0, 1, 0, 0, 0, 0, 0, 0, 0]\\
parental label & $\boldsymbol{\gamma}_i$ & [0, 1, 0, 1, 0, 0, 0, 0, 0, 0, .2]\\
length & $|\boldsymbol{x}_{i,j,k}|$ & 18 (tokens)\\
site ID & $i$ & 905\\
data split & & train \\
\hline
\end{tabular}
\caption{\label{sample}
An example of data sample.}
\end{table}

%\begin{figure}
%    \centering
%    \includegraphics[width=\linewidth]{distribution1}
%    \caption{The distribution of sentence lengths in the dataset for each data split and for independent test dataset SD.}
%    \label{distribution}
%\end{figure}

\paragraph{Dataset Example}
A data point concerning the WH site ``\emph{Kalwaria Zebrzydowska: the Mannerist Architectural and Park Landscape Complex and Pilgrimage Park}'' in Poland justified with Criteria (ii) and (iv) is shown in Table~\ref{sample}, with the attributes of text data $\boldsymbol{x}_{i,j,k}$, sentence label as discrete index $k$, sentence label as one-hot vector $\boldsymbol{y}_{i,j,k}$ (appended with 0 in the end for the class ``Others''), parental label as vector $\boldsymbol{\gamma}_i$ (appended with 0.2), sample length $|\boldsymbol{x}_{i,j,k}|$ in terms of the number of tokens, index of parental WH site $i$, and the data split.
%The distribution of sample length in terms of number of words in the sentence is shown in Figure~\ref{distribution}.
%The three data splits and the independent test dataset SD roughly have the same distribution in terms of sentence length.

\section{Proof of the Equivalence}
\label{proof}
Here we will show that the Vanilla Label Smoothing (LS) defined in Equations~\ref{eq_variant} and \ref{eq_softmax} is equivalent to the original LS assigning noise to all classes.

\begin{proof}
The LS defined in \citet{szegedy2016rethinking}:
\begin{equation}
    %y_k^{LS}=y_k(1-\alpha)+\alpha/K
    q'(k)=(1-\epsilon)\boldsymbol{\delta}_{k,y}+\frac{\epsilon}{K}
\end{equation}
could be rewritten as following to fit the context of mathematical notations in this paper:
\begin{equation}
\label{eq_origin}
    \boldsymbol{y}_{i,j,k}^{O}=(1-\epsilon)\boldsymbol{y}_{i,j,k}+\frac{\epsilon}{K}\boldsymbol{1},
\end{equation}
where $\boldsymbol{y}_{i,j,k}$ is a one-hot vector of ``ground-truth" label, $K$ is the total number of classes (instead of $\kappa+1$ in the paper for brevity and generality), $\epsilon$ is smoothing parameter as scalar, and $\boldsymbol{1}$ is a vector of 1s of size $K\times 1$.

On the other hand, the Vanilla LS proposed in this paper could be written as:
\begin{equation}
\label{eq_vanilla}
    \boldsymbol{y}_{i,j,k}^{V}=\mathbf{f}(\boldsymbol{y}_{i,j,k}+\alpha\boldsymbol{1})=\frac{e^{{\boldsymbol{y}_{i,j,k}}+\alpha\boldsymbol{1}}-\boldsymbol{1}}{{e^{({{\boldsymbol{y}_{i,j,k}}+\alpha\boldsymbol{1}})^T}}\boldsymbol{1}-K}.
\end{equation}

We will show that when
\begin{equation}
\label{eq_epsilon}
    \epsilon=\frac{(e^\alpha-1)K}{e^{1+\alpha}+(K-1) e^\alpha-K},
\end{equation} the vectors in Equations~\ref{eq_origin} and \ref{eq_vanilla} are the same.

First, it is trivial that both the vectors are with the same shape of $\boldsymbol{y}_{i,j,k}$, i.e., $K\times1$, and that the sums of all entries in both vectors are 1; e.g., observe that the denominator of the right-hand side of Equation \ref{eq_vanilla} is equal to the vectorised summation of the values of the nominator. 

Second, we assume, without loss of generality, that the ``ground-truth" of the one-hot vector $\boldsymbol{y}_{i,j,k}$ is at its first entry, which means that $\boldsymbol{y}_{i,j,k}=[1,0,...,0]_{K\times1}$.
Then both vectors could be rewritten as:
\begin{equation}
\label{eq_origin_en}
    \boldsymbol{y}_{i,j,k}^{O}=\left[1-\epsilon+\frac{\epsilon}{K},\frac{\epsilon}{K},...,\frac{\epsilon}{K}\right]_{K\times1},\\
\end{equation}
\begin{equation}
\label{eq_vanilla_en}
    \boldsymbol{y}_{i,j,k}^{V}=\left[\frac{e^{1+\alpha}-1}{S},\frac{e^{\alpha}-1}{S},...,\frac{e^{\alpha}-1}{S}\right]_{K\times1},\\
\end{equation}
where $S:=e^{1+\alpha}+(K-1)e^\alpha-K$.

Substituting Equation~\ref{eq_epsilon} into the entries in Equation~\ref{eq_origin_en}, the first entry could be rewritten as $1-\epsilon+\frac{\epsilon}{K}=1-\frac{(e^\alpha-1)K}{S}+\frac{e^\alpha-1}{S}=\frac{S-(e^\alpha-1)K+e^\alpha-1}{S}=\frac{e^{1+\alpha}+(K-1)e^\alpha-K-Ke^\alpha+K+e^\alpha-1}{S}=\frac{e^{1+\alpha}-1}{S}$.
And the other entries could be rewritten as $\frac{\epsilon}{K}=\frac{e^\alpha-1}{S}$.
Both types of entries are exactly the same as the ones shown in Equation~\ref{eq_vanilla_en}.

Last, we will show that $\epsilon$ has a one-to-one relation with $\alpha$ based on Equation~\ref{eq_epsilon} when $\alpha\geq0$.
The partial derivative of $\epsilon$ with respect to $\alpha$:
\begin{equation}
    \frac{\partial \epsilon}{\partial \alpha}=\frac{Ke^{\alpha}(e-1)}{(e^{1+\alpha}+(K-1)e^\alpha-K)^2}>0
\end{equation}
is non-negative, suggesting that the function is monotonic.
Furthermore, $\epsilon=0$ when $\alpha=0$, and $\lim\limits_{\alpha\to+\infty}\epsilon=\lim\limits_{\alpha\to+\infty}\frac{K}{\frac{e^\alpha(e-1)}{e^\alpha-1}+K}=\frac{K}{e-1+K}>0$ when $\alpha\to+\infty$, suggesting that it is incremental. 
This means that a unique $\epsilon\in\left[0,\frac{K}{e-1+K}\right)$ always exists for any non-negative $\alpha$ and \emph{vice versa}.
\end{proof}

\section{Model Implementation Detail}
\label{model_detail}

For all baselines, Adam \citep{kingma2017adam} is used as the optimizer with L2 regularization. 
Hyperparameter tuning is conducted as grid-search within a small range for each one being searched (and/or selected according to common experience if not mentioned), based on the top-k accuracy on validation split with an early-stopping criterion of 5 epochs, if not explicitly mentioned below.
The models are implemented in PyTorch \citep{Rao2019} and experiments are performed on NVIDIA Tesla P100 GPU (N-gram, GRU+Attn, BERT) and Intel Core i7-8850H CPU (BoE, ULMFiT), respectively.

\begin{figure*}
    \centering
    \includegraphics[width=0.9\linewidth]{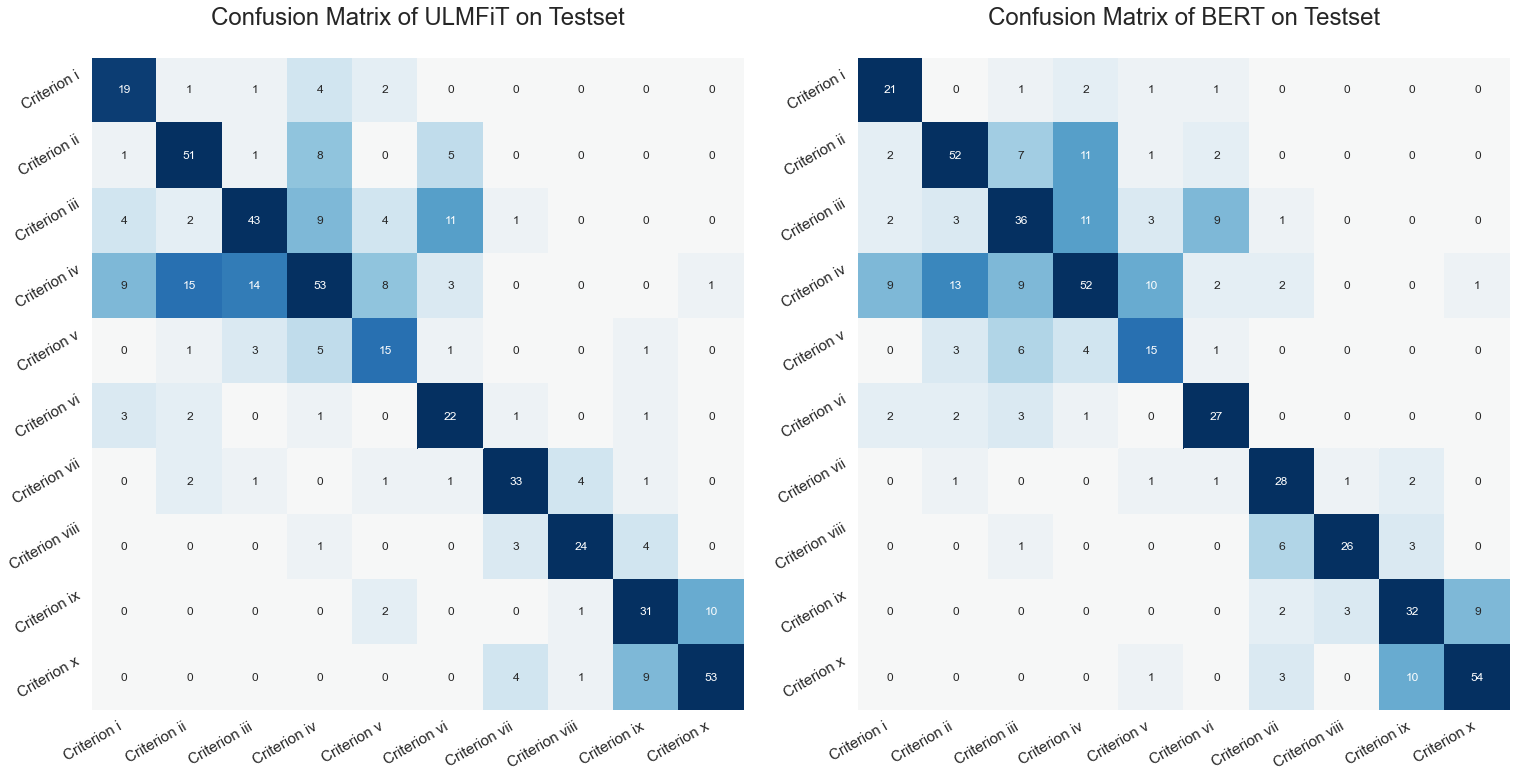}
    \caption{The confusion matrices of ULMFiT and BERT on test split.}
    \label{confusion}
\end{figure*}

\paragraph{N-gram}
The N-gram model used the TfidfVectorizer from Scikit-learn Python library %\citep{scikit-learn} 
to get an embedding vector of all 1-grams and 2-grams in the sample that appeared at least twice in the vocabulary.
The embedding vectors are then fed in a 2-layer Multi-layer Perceptron (MLP) to get the model prediction.

Hyperparameter tuning is performed on the size of the MLP hidden layer in \{50, 100, 150, 200\}, batch size in \{64, 128, 256\}, L2 in \{0, 1e-5. 1e-4\}, and dropout rate in \{0.1, 0.2, 0.5\} with 108 configurations.
The best configuration applied in later experiments of Label Smoothing (LS) has a hidden dimension of 200, batch size of 128, L2 of 1e-5, learning rate of 2e-4, and dropout rate of 0.5.

\paragraph{BoE}
The Bag-of-Embedding (BoE) model used the GloVe-6B-300d vectors\footnote{\href{https://nlp.stanford.edu/projects/glove/}{https://nlp.stanford.edu/projects/glove/}} as initial embeddings, which are set to be tunable during training. Only words that have a higher frequency than a threshold in the full dataset will be kept, while the others will be transformed to a special $<\mathrm{UNK}>$ token. The word embeddings of all words in the sentence is averaged before being fed to a 2-layer MLP.

Hyperparameter tuning is performed on the size of the MLP hidden layer in \{50, 100, 150, 200\}, batch size in \{64, 128, 256\}, and frequency threshold in \{1, 3, 5\} with 36 configurations.
The best model has a hidden dimension of 200, batch size of 64, cut-off frequency of 1, L2 of 1e-5, learning rate of 5e-4, and dropout rate of 0.1.

%\paragraph{CNN}
%The CNN model used the 300 dimensional GloVe embeddings. The sentence vector is fed in four successive 1-dimensional convolution layers of 128 channels with kernel size of 3, followed by ELU non-linearity and batch normalization. The output from CNN block goes through average pooling before fed to a 2-layer MLP.

%Hyper-parameter tuning is performed on whether or not the GloVe embeddings are frozen during training, the size of the MLP hidden layer, batch size, L2, learning rate, and dropout rate with 108 configurations.
%The best model has a unfrozen GloVe, hidden dimension of 128, batch size of 128, L2 of 1e-5, learning rate of 5e-4, and dropout rate of 0.5.

\begin{table*}
\centering
\small
\begin{tabular}{lrrrrrrr}
\hline \textbf{Performance} & \textbf{N-gram} & \textbf{BoE} & \textbf{GRU+Attn} & \textbf{ULMFiT} & \textbf{BERT}\\ \hline
\textbf{Infrastructure} & GPU & CPU & GPU & CPU & GPU$\times4$\\
\textbf{Training Time per Item} (s)& 0.34 & 0.18 & 0.03 & 2.53 & 0.54\\
\textbf{Training Time per Epoch} (s)& 12.69 & 3.18 & 1.97 & 213.61* & 46.20\\
\textbf{Early-Stopping Criteria} & 5 & 5 & 5 & 3 & 10\\
\textbf{Training Epochs} & 32 & 20 & 15 & 7** & 10\\
\textbf{Trainable Parameters} (M) & 3.82 & 1.88 & 0.18 & 24.55 & 109.49\\
\textbf{Inference Time per Item} (s) & 0.0031 & 0.0007 & 0.2245 & 0.0589 & 0.5542 \\
\textbf{Inference Time for SD} (s) & 6.92 & 1.44 & 4.44 & 151.75 & 1598.06\\
\hline
\multicolumn{2}{l}{\small *1180.20 during language model fine-tuning.}\\
\multicolumn{2}{l}{\small **11 during language model fine-tuning.}
\end{tabular}
%\small
%*1180.20 during language model fine-tuning
\caption{\label{performance}
The model performance in terms of resource occupancy and inference time. The inference is conducted on Intel Core i7-8850H CPU. \emph{Inference time per Item} shows the average time the model uses to make a prediction on one sentence. And \emph{Inference time for SD} shows the total time the model needs to fully process and predict the independent Short Description (SD) test set.}
\end{table*}

\paragraph{GRU+Attn}
The GRU+Attn model also used the GloVe-6B-300d as embeddings, which are frozen during the training. The embedding sequence is then fed into a GRU network. 
Word-level attention \citep{yang2016hierarchical} is applied to compute the sentence vector by a learned word context vector and the last hidden state of the GRU.
The sentence vector is fed to a 1-layer feed-forward network for the output of the model.

Hyperparameter tuning is performed on the size of the hidden layer in GRU in \{64, 128, 256\}, whether or not to use bi-directional GRU, batch size in \{64, 128, 256\}, L2 in \{0, 1e-5, 1e-4\}, learning rate in \{1e-3. 5e-4. 2e-4\}, and dropout rate in \{0, 0.1, 0.2, 0.5\} with 648 configurations.
The best model is a uni-dimensional GRU with hidden dimension of 128, batch size of 256, L2 of 1e-5, learning rate of 1e-3, and dropout rate of 0.1.

\paragraph{ULMFiT}
The ULMFiT model employs the idea of Universal Language Model Fine-tuning from a general-domain pretrained language model on Wikitext-103 with AWD-LSTM architecture \citep{ulmfit}.
A domain-specific language model is then fine-tuned with the full UNESCO WHL dataset including SD using fastai API \citep{Howard2020}.
One epoch is trained with a learning rate of 1e-2, with only the last layer unfrozen, reaching a perplexity of 46.71.
Then the entire model is unfrozen and further trained for 10 epochs, with a learning rate of 1e-3, obtaining a fine-tuned WH domain-specific language model reaching a 30.78 perplexity.
Some examples of the language model at this step are shown here, starting with the given phrases marked in bold:
\begin{quote}
\small
    \emph{\textbf{This site is unique because} it is the only example of a complex of karst complexes that is clearly recognised as being of outstanding universal value. The island of zanzibar has been inscribed as a world heritage site in <num>. The inscriptions, which bear witness to the civilisation of...}
    
    \emph{\textbf{This architecture has a special layout}, especially in the form of the body of the building. The planet's primary feature is the addition of the ideal island, which lies at an elevation of <num>m above the sea floor, and is home to some <num>...}
\end{quote}

The encoder of the fine-tuned language model is loaded in PyTorch followed by a Pooling Linear Classifier\footnote{\href{https://fastai1.fast.ai/text.models.html}{https://fastai1.fast.ai/text.models.html}} for classifier fine-tuning.
Gradual unfreezing is applied in a simplified manner to prevent catastrophic forgetting:
1) for the 1st epoch, only the decoder is unfrozen and trained with a learning rate of 2e-2;
2) for the 2nd to 4th epoch, one more layer is unfrozen each time and trained with a learning rate of 1e-2, 1e-3, and 1e-4, respectively;
3) from the 5th epoch onward, the full model is unfrozen and trained with a learning rate of 2e-5.
An early-stopping criterion of 3 is applied.

No extensive hyperparameter tuning is performed since: 1) tuning ULMFiT is expensive on CPU; 2) the hyperparameter configuration from experience suggested by \citet{Howard2020} and \citet{ulmfit} already performs reasonably well; 3) the purpose of this study is not necessarily finding the best hyperparameter.
The final model uses batch size of 64, L2 of 1e-5, and the default dropout rate for the decoder.

\paragraph{BERT}
The BERT model uses the uncased base model %pretrained on large corpus \citep{devlin-etal-2019-bert} 
using The Transformers library \citep{wolf2020transformers}.
The pooler output processed from the last hidden-state of the [CLS] token during pretraining is fed into a 1-layer feed-forward network to fine-tune the classifier \citep{Sun2019}.
An early-stopping criterion of 10 is applied.

Hyperparameter tuning is performed on the batch size in \{16, 24, 48, 64\}, L2 in \{0, 1e-5, 1e-4\}, and dropout rate in \{0, 0.1, 0.2\} with 36 configurations.
The best model uses batch size of 64, L2 of 1e-4, learning rate of 2e-5, and dropout rate of 0.2.

\paragraph{LS Configuration Tuning}
A single random seed 1337 is used for hyperparameter tuning.
Afterwards, ten random seeds in \{0, 1, 2, 42, 100, 233, 1024, 1337, 2333, 4399\} are used to tune the LS configuration with $\alpha\in\{0,0.01,0.05,0.1,0.2,0.5,1\}$ for all three variants.
The best LS configuration is selected based on the sum of the lower bound of 95\% confidence interval on both top-1 and top-k accuracy.
The best LS configuration is then used to evaluate the model performance on single seed 1337.
The total runs on each baseline are, therefore, the sum of the number of hyperparameter configurations and random seeds experiments (which is 210).

\section{Extended Model Performance}
\label{model_performance}
\paragraph{Resource and Time}
Table~\ref{performance} shows some further information on the model performance in terms of training resource utilization, model size, and inference time.
Training processes are conducted on CPU or GPU, respectively, while inference is fully conducted with CPU.

It can be noted that the best-performing models ULMFiT and BERT also consume the most resources, in terms of training time and infrastructure usage, and have the largest model sizes.
Though most time-consuming during training, ULMFiT takes a remarkably short time for inference on CPU compared to BERT.
This suggests that ULMFiT might be an optimal choice for further development and application when time is a critical matter.

\begin{figure*}
    \centering
    \includegraphics[width=0.85\linewidth]{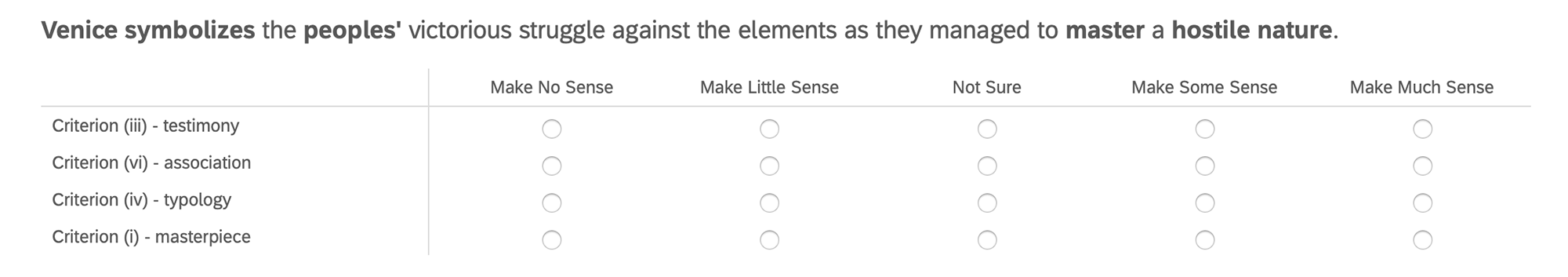}
    \caption{The evaluation interface on Qualtrics.}
    \label{interface}
\end{figure*}

\paragraph{Confusion Matrices}
The confusion matrices of the best-performing ULMFiT and BERT models on the test split are shown in Figure~\ref{confusion}.
It can be seen that certain criteria are easily confused as the others, such as sentences with a ``ground-truth'' label of criterion (iv) can be confused as criteria (i), (ii), and (iii), and \emph{vice versa};
while criterion (iii) might be confused easily as criterion (vi), but \textbf{NOT} \emph{vice versa}.
This complex association relationship is extensively discussed in \citet{bai2021}.
%Table~\ref{post-hoc} shows the result of post hoc comparisons using the Tukey HSD test for the one-way ANOVA conducted on the F1 metric of cultural and natural OUV classes by all models.
%Only intra-category pairs of OUV criteria that are significantly different with each other are shown. 

%\begin{table}
%\centering
%\small
%\begin{tabular}{llrr}
%\hline \textbf{Class 1} & \textbf{Class 2} & \textbf{Mean Diff} & %\textbf{$p$ value} \\ \hline
%\multicolumn{3}{l}{\small\textbf{Cultural OUV}}\\
%C1 & C2 & -11.38** & .006\\
%C1 & C4 & 11.17** & .007\\
%C2 & C4 & 10.32* & .015\\
%C2 & C5 & 22.55*** & .001\\
%C3 & C5 & 16.00*** & .001\\
%C4 & C5 & 12.23** & .002\\
%C5 & C6 & -16.26*** & .001\\ \hline \hline
%\multicolumn{3}{l}{\small\textbf{Natural OUV}}\\
%N7 & N9 & 10.90*** & .001\\
%N7 & N10 & -8.12** & .003\\
%N8 & N9 & 8.94** & .001\\
%N8 & N10 & -10.08*** & .001\\
%N9 & N10 & -19.03*** & .001\\
%\hline
%\multicolumn{4}{l}{\small *$p<0.05$.}\\
%\multicolumn{4}{l}{\small **$p<0.01$.}\\
%\multicolumn{4}{l}{\small ***$p<0.001$.}\\
%\end{tabular}
%\small
%*1180.20 during language model fine-tuning
%\caption{\label{post-hoc}
%The multiple comparison of means using Tukey HSD on the F1 metric of OUV classes by all models on both validation and test splits within both cultural and natural categories.}
%\end{table}

\section{Expert Evaluation Details}
\label{expert_evaluation}

\paragraph{Materials}
The materials about the WH site ``\emph{Venice and Its Lagoon}'' for expert evaluation were harvested from three data sources:
1) all 14 sentences from \textbf{Justification for Criteria} section of Statements of OUV (SOUV), where each sentence has one ``ground-truth'' sentence label and a parental site label of Venice, which is also within the data $\boldsymbol{X}_i$ used during model training and testing;
2) all 13 sentences from \textbf{Brief Synthesis} section of SOUV, where sentences only have the same multi-label parental label of Venice, which is similar with the SD test data $\boldsymbol{S}_i$ used for generalization test;
 3) \textbf{Social Media} data sampled from a total of 1687 social media posts where a textual description is written, collected from Flickr in the region of Venice with a resolution of 5km using Flickr API\footnote{\href{https://pypi.org/project/flickrapi/}{https://pypi.org/project/flickrapi/}}.
%A similar data pre-processing as described in the main body was performed for all three data sources, creating sentence-level textual data.
Among the 1687 social media posts, there are 820 unique textual descriptions in English.
By splitting the unique posts into sentences, removing HTML symbols, and filtering out the texts about camera parameters, image formats, and advertisements, 1132 sentences were obtained.
The 1132 sentences were fed into the trained BERT and ULMFiT models.
The sentences were further filtered based on the predictions:
1) the total confidence scores of top-3 predictions need to be larger than 0.8 by both models;
2) the Intersection over Union of top-3 predictions by two models needs to be larger than 0.5 (i.e., maximum one different predicted class).
As a result, 388 Social Media sentences that potentially convey OUV-related information were obtained.
Furthermore, 29 sentences were randomly sampled from those 388 for the expert evaluation.

\begin{table*}
\centering
\small
\begin{tabular}{cllrrrrrrrr}
\hline \textbf{Data Source} & \textbf{Type-1}  & \textbf{Type-2} & \textbf{$M_1$} & \textbf{$M_2$} & \textbf{$n_1$} & \textbf{$n_2$} & \textbf{$U$ value} & \textbf{$p$ value}\\ \hline \hline
\textbf{Justification} & top-1 prediction & top-3 prediction & 5 & 2 & 120 & 240 & 8157.0*** & <0.001\\
\textbf{of Criteria} & top-1 prediction & negative class & 5 & 2 & 120 & 120 & 3161.0*** & <0.001\\
& top-3 prediction & negative class & 2 & 2 & 240 & 120 & 12638.0* & 0.026\\ \hline \hline
\textbf{Brief} & top-1 prediction & top-3 prediction & 4 & 2 & 96 & 192 & 6256.0*** & <0.001\\
\textbf{Synthesis} & top-1 prediction & negative class & 4 & 2 & 96 & 96 & 2401.5*** & <0.001\\
& top-3 prediction & negative class & 2 & 2 & 192 & 96 & 7603.5** & 0.006\\ \hline \hline
\textbf{Social}& top-1 prediction & top-3 prediction & 3 & 2 & 232 & 464 & 40629.0*** & <0.001\\
\textbf{Media}& top-1 prediction & negative class & 2 & 1 & 232 & 232 & 13784.5*** & <0.001\\
& top-3 prediction & negative class & 2 & 1 & 464 & 232 & 39284.5*** & <0.001\\
\hline
\multicolumn{5}{l}{\small *$p<0.05$, **$p<0.01$, ***$p<0.001$.}\\
%\multicolumn{5}{l}{\small **$p<0.01$.}\\
%\multicolumn{5}{l}{\small ***$p<0.001$.}\\
\end{tabular}
%\small
%*1180.20 during language model fine-tuning
\caption{\label{u-test}
The results of post-hoc Mann-Whitney $U$ tests for the three types of labels within each data source. The medians ($M$) and counts ($n$) of each type are given together with the statistics from $U$ tests.}
\end{table*}

\begin{table*}
\centering
\small
\begin{tabular}{lclrrrr}
\hline \textbf{Text} & \textbf{Criteria} & \textbf{Source}  & \textbf{Type} & \textbf{BERT} & \textbf{ULMFiT} & \textbf{Expert Ratings} \\ \hline
\emph{With the unusualness of an archaeological}\\
\emph{site which still breathes life, Venice bears}& iii & justification & top-1 & 0.744 & 0.825 & 5,5,5,3,5,5,4,5\\
\emph{testimony unto itself}.\\
\emph{Human interventions show high technical}\\
\emph{and creative skills in the realization of the} & i & synthesis & top-1 & 0.607 & 0.590 & 4,5,5,1,4,4,2,5\\
\emph{hydraulic and architectural works in the}\\
\emph{lagoon area.}\\
\emph{In 1952, the station was finalized on a}\\
\emph{design by the architect Paul Perilli.} & i & social media & top-1 & 0.757 & 0.529 & 5,4,1,1,1,3,1,1\\
\hline
\end{tabular}
%\small
%*1180.20 during language model fine-tuning
\caption{\label{example}
Some example ratings on sentence-criterion relevance by human experts. The confidence scores by the computer models BERT and ULMFiT are also given.}
\end{table*}

\paragraph{Survey Design}
Each of the 56 sentences was fed into BERT and ULMFiT models to obtain the predictions and confidence scores.
The predicted selection criteria with the highest confidence scores by both models were considered as the \textbf{top-1 predictions}.
Two other criteria within the top-3 classes predicted by both models with relatively high confidence scores were considered as the \textbf{top-3 predictions} for the survey.
Another random cultural criterion that was not predicted by any model to be top-3 classes was considered as the \textbf{negative class} for each sentence.
Criteria for natural heritage were not sampled as negative classes as they are not easily confused with the positive cultural ones.
As a result, each sentence got \textbf{four} criteria to be evaluated.
All four criteria were presented in a random order for each sentence, asking for an evaluation about the relevance of the sentence conveying the criterion on a 5-point Likert scale (from ``\emph{5: make much sense}'', to ``\emph{1: make no sense}'').
The ``important'' words with higher attention weights in the GRU+Attn model were highlighted in bold.
An example of such evaluation on Qualtrics platform is shown in Figure~\ref{interface}.
The sentences from the three data sources were grouped in four separate sessions, while the social media data were split into two sessions.
The session of ``\emph{justification for criteria}'' were always presented first during evaluation, also as a practice for the experts.
The other three sessions were presented in a randomized order to prevent systematic errors caused by impatience or tiredness.
Additional questions about the familiarity for heritage value identification, familiarity about Venice, confidence of evaluation, usefulness of highlighted words, and overall enjoyment and difficulty of the exercise were respectively raised before and after the evaluation, also with 5-point Likert scale.
Note the number of samples involved in the in-depth expert evaluation is relatively small, which is not uncommon in qualitative validation.
Moreover, we plan to conduct online non-expert human evaluation in follow-up studies, which could involve more participants with larger sample sentences.
It would, however, serve a different purpose than the expert evaluation presented.

\paragraph{General Analyses}
The evaluations took $55.10\pm20.74$ minutes to finish.
The eight experts are all very familiar with the concept of OUV ($4.38\pm0.70$) and the heritage values and attributes identification ($4.75\pm0.43$), while not all are familiar with OUV justification ($3.00\pm1.50$), nor with the cultural heritage in Venice ($3.00\pm1.41$).
The experts agree that the exercise in the evaluation was very hard ($4.13\pm0.93$) and not so enjoyable ($2.63\pm1.32$).
They are more confident with identifying irrelevant sentence-criterion pairs ($3.88\pm0.78$) than evaluating the relevant ones ($3.00\pm1.12$).
These show that the results of the expert evaluation are sufficiently reliable, that the heritage experts are cautious and critical of the process, that OUV justification is a difficult task even for experts as it is time-consuming and knowledge-demanding, and that a computational model is urgently needed to automate the classification if to be applied with massive social media data.
The experts are not fully convinced that the highlighted words helped them with the justification process ($2.88\pm1.05$), since the words provide both relevant information ($3.13\pm1.27$) and irrelevant information ($4.38\pm0.70$).
This suggests that the explainability of the model using GRU+Attn attention mechanism needs further development.

\paragraph{Evaluation Results}
Since the expert evaluations are in ordinal scales, non-parametric statistical tests, including Kruskal-Wallis $H$ tests (analogous to ANOVA) and Mann-Whitney $U$ tests (analogous to $t-$ test), are conducted.
The statistic analyses 
%including independent $t$-test, one-way ANOVA, and post hoc comparisons
are performed with Scipy\footnote{\href{https://docs.scipy.org/doc/scipy/reference/stats.html}{https://docs.scipy.org/doc/scipy/reference/stats.html}} and Statsmodels\footnote{\href{https://github.com/statsmodels/statsmodels}{https://github.com/statsmodels/statsmodels}} libraries.
Kruskal-Wallis $H$ tests show significant differences 
%among the three data sources [$H(2)=110.212, p<0.001$], and [$H(2)=187.757, p<0.001$].
among the three types of criteria labels for all data sources, including for ``\emph{justification of criteria}'' [$H(2)=68.412, p<0.001$], for ``\emph{brief synthesis}'' [$H(2)=40.351, p<0.001$], and for ``\emph{social media}'' [$H(2)=102.321, p<0.001$].
Post-hoc Mann-Whitney tests were used to compare all pairs of groups, as is shown in Table~\ref{u-test}.
The all-significant results of $U$ tests show that the human experts gave significantly higher ratings to top-1 predictions than top-3 predictions, and to top-3 predictions than negative classes.
In other words, the human experts and computer models are consistently similar in differentiating the positive and negative criteria for the sentences concerning their relevance.
Some exemplary ratings of the experts and model predictions are given in Table~\ref{example}.
It shows that the opinion of experts easily diverge, that some experts seem to be rather cautious during evaluation and rate lower for the social media data, and that it is difficult even for human experts to reach an agreement without further discussion.

\end{document}